\documentclass{article}

\usepackage[preprint]{neurips_2026}

\usepackage[utf8]{inputenc}
\usepackage[T1]{fontenc}
\usepackage{hyperref}
\usepackage{url}
\usepackage{booktabs}
\usepackage{amsfonts}
\usepackage{amssymb}
\usepackage{amsmath}
\usepackage{amsthm}
\usepackage{nicefrac}
\usepackage{microtype}
\usepackage{xcolor}
\usepackage{colortbl}
\usepackage{algorithm}
\usepackage{algorithmic}
\usepackage{graphicx}
\usepackage{multirow}
\usepackage{subcaption}
\usepackage{placeins}
\usepackage{float}
\usepackage{marvosym}
\usepackage{pgfplots}
\pgfplotsset{compat=1.18}

\newcommand{\bW}{\mathbf{W}}
\newcommand{\bH}{\mathbf{H}}
\newcommand{\bg}{\mathbf{g}}
\newcommand{\bx}{\mathbf{x}}
\newcommand{\by}{\mathbf{y}}

\newcommand{\bF}{\mathbf{F}}
\newcommand{\cL}{\mathcal{L}}
\newcommand{\cD}{\mathcal{D}}
\newcommand{\RR}{\mathbb{R}}
\newcommand{\EE}{\mathbb{E}}

\newtheorem{theorem}{Theorem}[section]
\newtheorem{proposition}[theorem]{Proposition}

\newtheorem{lemma}[theorem]{Lemma}

\title{REAL-Q: E2E LLM Quantization via Dynamic Gradient Descent}

\author{
 \textbf{Qian Zhang\textsuperscript{1$*$}},
 \textbf{Yaoming Li\textsuperscript{1$*$}},
 \textbf{Zhewen Tan\textsuperscript{1}},
 \textbf{Yanshu Wang\textsuperscript{1}},
 \textbf{Heng Lu\textsuperscript{1}},
 \textbf{Kun Su\textsuperscript{2}},\\
 \textbf{Zongwei Lv\textsuperscript{1}},
 \textbf{Wenhan Yu\textsuperscript{1}},
 \textbf{Yongge Ma\textsuperscript{1}},
 \textbf{Yinjun Han\textsuperscript{3}},
 \textbf{Ruikang Liu\textsuperscript{3}},
 \textbf{Tong Yang\textsuperscript{1\Letter}}
\\
{
\shortstack[c]{
  \textsuperscript{1} Peking University 
  \textsuperscript{2} Northeastern University 
  \textsuperscript{3} ZTE Corporation
}
}\\
{\small 
  \textsuperscript{$*$}Equal contribution \quad
  \textsuperscript{\Letter} Correspondence:\href{mailto:yangtong@pku.edu.cn}{yangtong@pku.edu.cn}
}
} 

\begin{document}

\maketitle

\begin{abstract}
Post-training quantization (PTQ) is essential for deploying large language models (LLMs) under strict resource constraints. State-of-the-art PTQ methods quantize each layer with a single closed-form second-order solver: to remain analytically tractable, they heavily approximate the global loss (dropping cross-channel coupling, pooling output rows into groups), and they then freeze the resulting Hessian across the entire layer, with no way to refresh it as the loss landscape shifts column by column---a phenomenon we call \textit{information misalignment}. We propose \textbf{REAL-Q} (\textbf{R}eal-time \textbf{E}2E-loss \textbf{A}ligned \textbf{L}LM \textbf{Q}uantization), a novel PTQ paradigm that breaks this compromise: instead of diluting the objective for the sake of analytic tractability, REAL-Q targets an end-to-end-aligned surrogate of the global loss and refines it via fine-grained, dynamic Block-wise Gradient Descent applied after every column block (128 columns). By coupling this fine-grained correction with a sliding window mechanism for smooth cross-layer transitions, REAL-Q effectively mitigates error propagation across the network. On LLaMA-3.1 (8B and 70B) and Qwen3 (0.6B--32B) at W4A16, REAL-Q reduces end-to-end KL divergence by up to $\sim$49\% relative to state-of-the-art globally-guided methods.
\end{abstract}

\section{Introduction}
\label{sec:intro}

Large language models (LLMs) have demonstrated remarkable capabilities across a wide range of natural language processing tasks \citep{grattafiori2024llama}. However, deploying models with billions of parameters incurs substantial compute and memory costs. Post-training quantization (PTQ) offers a practical solution by compressing model weights to low-bit representations without requiring costly retraining \citep{gholami2021survey}.

Among PTQ methods, GPTQ \citep{frantar2023gptq} has emerged as the foundational algorithm for weight-only compression. Building upon the Optimal Brain Surgeon (OBS) framework \citep{hassibi1993optimal}, its column-wise solver has been widely integrated into mainstream inference engines (e.g., vLLM, TensorRT-LLM) and extended by a vast lineage of state-of-the-art techniques \citep{dettmers2024spqr,li2025gptaq,kim2025guidedquant}. Despite its success, GPTQ has a fundamental limitation:

\textbf{Inaccuracy of the Layer-Local Objective.} GPTQ minimizes a layer-wise reconstruction MSE $\|\bW\bx-\hat{\bW}\bx\|_2^2$, where $\bx$ is the full-precision activation. This local objective suffers from a dual misalignment. Upstream, it assumes full-precision input activations $\bx$ and ignores the perturbed inputs $\hat{\bx}$ arriving from previously quantized layers, leaving compounding errors uncorrected. Downstream, it weights all output-channel perturbations uniformly, ignoring that the true KL divergence depends on a strongly non-linear, highly coupled mapping through subsequent transformer blocks and the LM head. A surrogate that relies on unperturbed inputs and uniformly weights squared errors is therefore misaligned with the true quantization loss.

Recent works attempt to address these blind spots: GPTAQ \citep{li2025gptaq} recalibrates inputs to correct upstream errors, yet its per-layer loss remains a uniform reconstruction MSE. Conversely, GuidedQuant \citep{kim2025guidedquant} incorporates end-to-end gradients to capture downstream sensitivity. However, it still leaves upstream errors uncorrected, and it relies on structural approximations such as ignoring fine-grained cross-channel coupling and sharing the Hessian across output groups. In addition, because the true end-to-end Hessian is prohibitively expensive to recompute, the already-approximated matrix remains static within each layer. While the true loss landscape shifts as weights are progressively quantized column by column, the misaligned Hessian causes compensation errors to accumulate as the column-wise sweep progresses. We refer to this phenomenon as \textit{information misalignment}. Here, "information" refers to first- or second-order information such as the gradient and Hessian.

\textbf{The shared limitation of these methods is structural:} requiring the column-wise update to remain analytically solvable forces them to (i) heavily approximate the global loss---dropping cross-channel coupling, pooling output rows into groups, and otherwise compressing the second-order information until it fits a closed-form solver---and (ii) freeze the resulting Hessian over the entire layer, with no way to refresh it at finer granularity as the loss landscape shifts with each column quantization step. REAL-Q (\textbf{R}eal-time \textbf{E}2E-loss \textbf{A}ligned \textbf{L}LM \textbf{Q}uantization) addresses both limitations by interleaving \emph{fine-grained, dynamic} corrections into the column-wise sweep, via two components:

\textbf{An end-to-end-aligned objective: Aggregated Fisher MSE.} REAL-Q's surrogate is a transformer-block-output Fisher MSE, derived from a second-order Taylor expansion of the global KL divergence at the current block's output. Using the \emph{full} aggregated Fisher matrix preserves the cross-channel coupling that grouped baselines must discard, while the choice of expansion point keeps backpropagation confined to at most two adjacent blocks. A loss sliding window smooths the objective across block boundaries to prevent discontinuous jumps in the optimization landscape.

\textbf{Fine-grained, dynamic Block-wise Gradient Descent.} Rather than committing the entire layer to one frozen second-order solver, REAL-Q interleaves a single Adam \citep{kingma2014adam} gradient step into the GPTQ sweep \emph{after every column block} (128 columns). This raises the correction granularity from the per-layer level used by prior solvers to the column-block level---roughly two orders of magnitude finer---and lets each step react to the most recent quantized state, since the gradient is recomputed online against the actual partially quantized weights. Adam's diagonal preconditioner additionally adapts the per-coordinate scale on the fly, providing implicit second-order information without ever inverting a Hessian.

\begin{figure}[!t]
    \centering
    \includegraphics[width=\linewidth]{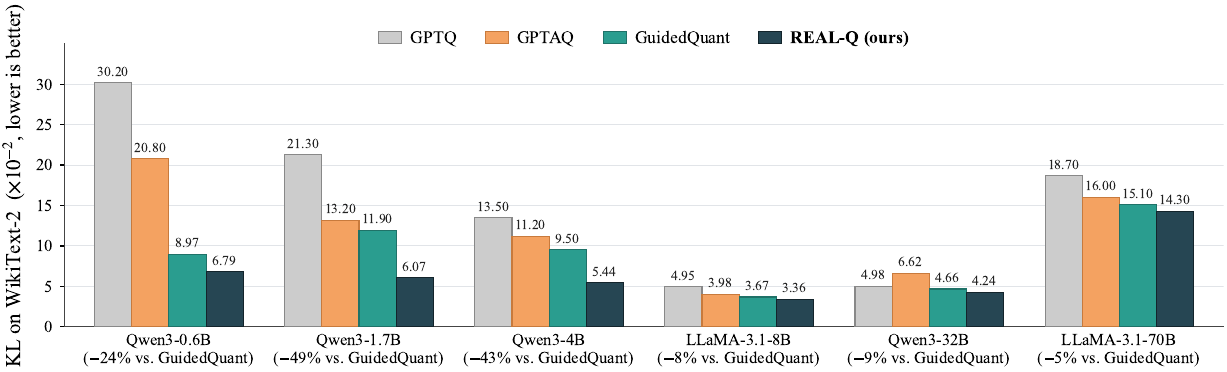}
    \caption{\textbf{Headline result.} KL divergence ($\times 10^{-2}$, lower is better) on WikiText-2 (W4A16, per-row). REAL-Q substantially outperforms GuidedQuant, the strongest prior baseline, with exact relative reductions annotated per model. Full seven-model results are in Appendix~\ref{app:exp_full}.}
    \label{fig:headline}
\end{figure}

In summary, our main contributions are as follows:

\begin{itemize}
\item \textbf{Aggregated Fisher MSE with Loss Sliding Window:} Within the GPTQ-lineage sequential pipeline, we adopt an end-to-end-aligned objective in place of the layer-local MSE. We formulate an aggregated Fisher MSE that preserves cross-channel coupling while keeping the backpropagation overhead confined to localized block boundaries. To stabilize the sequential quantization process, we introduce a \textbf{loss sliding window} mechanism that smooths the transition of the objective across consecutive transformer blocks and prevents discontinuous jumps in the loss landscape.
\item \textbf{Dynamic Block-wise Gradient Descent (Block-GD):} We address the \textit{information misalignment} of prior works by replacing static analytical solvers with a dynamic first-order approach. By aligning Adam-driven gradient feedback directly with column-block updates, this mechanism adapts to the shifting loss landscape and corrects accumulated upstream errors online.
\item \textbf{Empirical Performance:} Experiments on LLaMA-3.1 (8B and 70B) and Qwen3 (0.6B--32B) show that REAL-Q achieves the lowest KL on every model evaluated, reducing KL by up to $\sim$49\% on Qwen3-1.7B at W4A16 (Figure~\ref{fig:headline}).
\end{itemize}

\section{Related Work}
\label{sec:related}

\paragraph{Second-order post-training quantization.}
The Optimal Brain Damage (OBD) \citep{lecun1989optimal} and Optimal Brain Surgeon (OBS) \citep{hassibi1993optimal} frameworks established the foundation for using second-order Hessian information in network compression. GPTQ \citep{frantar2023gptq} successfully scaled OBS to billion-parameter LLMs. By processing weight rows concurrently and employing a shared, static inverse Hessian to compensate for quantization errors column by column, GPTQ established the dominant pipeline for modern sequential PTQ.

\paragraph{Global-aware extensions of GPTQ.}
Recent advancements attempt to enhance the standard sequential GPTQ pipeline by incorporating broader model context into the quantization process. 
\textbf{GPTAQ} \citep{li2025gptaq} addresses upstream error accumulation by iteratively recalibrating the input activations. Instead of relying solely on fixed, full-precision inputs, it dynamically computes calibration activations using the outputs from previously quantized layers, thereby incorporating upstream quantization drift into the layer-wise reconstruction objective. 
\textbf{GuidedQuant} \citep{kim2025guidedquant} incorporates end-to-end downstream sensitivity into the local solver. It performs an initial backward pass over the calibration set to obtain per-token saliency scores, which are subsequently used to weight the input covariance matrix. To maintain the computational tractability of the analytical second-order update, GuidedQuant partitions the weight matrix rows into groups and computes a shared, static approximated Hessian for each group to drive the column-wise quantization.

\paragraph{Other related directions.} 
Extended discussions on rotation pre-processing (e.g., QuaRot \citep{ashkboos2024quarot}, which we adopt), weight-only representations, quantization-aware training, and classical Fisher applications are deferred to Appendix~\ref{app:related_extra}. This appendix also provides a detailed comparison with BRECQ \citep{li2021brecq}; while BRECQ shares conceptual similarities in block-level loss modeling, its overall quantization pipeline is fundamentally different from ours.

\section{Preliminaries}
\label{sec:prelim}

\subsection{Problem Formulation}

Consider a pre-trained LLM with $L$ transformer blocks, where each block $\ell$ contains linear modules (e.g., $\bW^{q}, \bW^{k}, \bW^{v}, \bW^{o}$ for attention and $\bW^{\text{up}}, \bW^{\text{gate}}, \bW^{\text{down}}$ for the MLP). For a linear module with weight matrix $\bW \in \RR^{m \times n}$, the goal of PTQ is to find a quantized weight $\hat{\bW}$ that minimizes the degradation in model output quality.

Given a calibration dataset $\cD = \{\bx_1, \ldots, \bx_N\}$, we measure quantization quality via the KL divergence between the original and quantized model outputs:
\begin{equation}
    \cL_{\text{KL}} = \EE_{\bx \sim \cD} \left[ D_{\text{KL}} \left( p_{\bW}(\cdot \mid \bx) \;\|\; p_{\hat{\bW}}(\cdot \mid \bx) \right) \right].
    \label{eq:kl_loss}
\end{equation}

Following recent studies \citep{kong2026kllens}, we adopt KL divergence as our primary evaluation metric, as it captures the full output distributional shift and correlates more faithfully with quantization-induced behavioral degradation than perplexity. For comparability with prior works, we also report perplexity and zero-shot task accuracy as secondary metrics.

\subsection{GPTQ Recap}
\label{sec:gptq_recap}
GPTQ \citep{frantar2023gptq} processes linear modules by quantizing weights column by column. At step $j$, GPTQ quantizes column $j$ to $\hat{\mathbf{w}}_j$, computes the residual $\boldsymbol{\delta}_j = \mathbf{w}_j - \hat{\mathbf{w}}_j$, and optimally updates the remaining unquantized columns to compensate: 
\begin{equation}
    \bW_{:, j+1:n} \leftarrow \bW_{:, j+1:n} - \frac{1}{[\bH^{-1}]_{jj}} \boldsymbol{\delta}_j \bH^{-1}_{j, j+1:n},
    \label{eq:gptq_update}
\end{equation}
where $\bH = \frac{2}{T} \sum_{t=1}^{T} \bx_t \bx_t^\top$ is the Hessian of the localized reconstruction MSE. To accelerate computation, GPTQ groups columns into blocks of size $B$. After applying Eq.~\ref{eq:gptq_update} within the block, a batch update compensates all trailing columns globally: $\bW_{:, b_{\text{end}}:n} \leftarrow \bW_{:, b_{\text{end}}:n} - \mathbf{E}_{\text{b}} \cdot \bH^{-1}_{b, b_{\text{end}}:n}$ ($b_{\text{end}}$ denotes the end of block $b$), where $\mathbf{E}_{\text{b}}$ is the accumulated block error.

\subsection{GuidedQuant: Saliency-Weighted Hessian}
\label{sec:guidedquant_recap}
While standard GPTQ relies on a uniform layer-wise reconstruction objective, GuidedQuant \citep{kim2025guidedquant} improves this by aligning the Hessian with the end-to-end loss. By performing an initial end-to-end backward pass over the calibration set, it obtains per-token saliency scores $s_t^{(\ell)} = \|\partial \cL / \partial \by_t^{(\ell)}\|^2$, which are then used to weight the GPTQ Hessian computation. To manage computational overhead, the weight matrix rows are partitioned into $N_g$ groups (we use $N_g = 4$ in all experiments), and a separate Hessian is maintained for each group:
\begin{equation}
    \bH^{(\ell)}_g = \sum_{t=1}^{T} s_{t,g}^{(\ell)} \cdot \bx_t \bx_t^\top, \quad g = 1, \ldots, N_g,
    \label{eq:saliency_hessian}
\end{equation}
where $s_{t,g}^{(\ell)}$ is the saliency score for token $t$ restricted to the output dimensions in group $g$. While computationally feasible, this group-sharing strategy inherently discards exact cross-channel coupling and degrades the precision of the second-order matrix.

\section{REAL-Q}
\label{sec:method}

\subsection{Overview}

\textbf{REAL-Q} augments the saliency-weighted PTQ pipeline with a coarse-to-fine optimization hierarchy. We retain the traditional column-wise analytical compensation as a preliminary coarse update, but augment the optimization trajectory with: (1) an \textbf{Aggregated Fisher-Weighted MSE} providing end-to-end gradient guidance; (2) \textbf{Dynamic Block-GD}, an active first-order correction applied immediately post-block to reduce residual errors (e.g., upstream perturbations and structural imprecision) missed by static solvers; and (3) a \textbf{Loss Sliding Window} to smoothly interpolate objectives across consecutive transformer blocks. Together, these components correct residual errors left by static second-order updates. Figure~\ref{fig:blockgd} presents a comparison between prior methods and REAL-Q. The complete pseudocode is provided in Appendix~\ref{app:algorithm}.

\begin{figure}[h]
    \centering
    \includegraphics[width=0.8\linewidth]{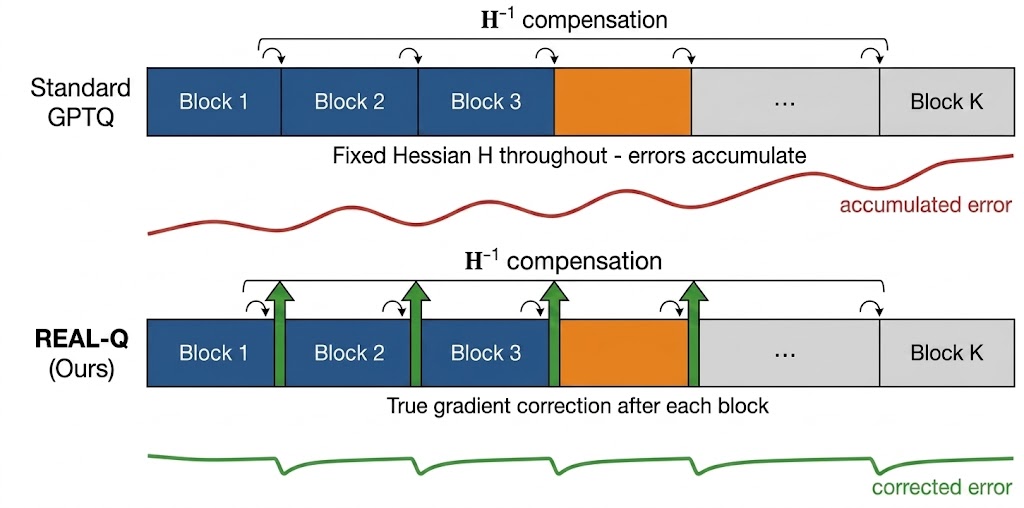}
    \caption{Comparison of static analytical quantization and REAL-Q. In prior static methods (e.g., standard GPTQ or GuidedQuant), a fixed Hessian approximation is used throughout, allowing quantization errors to accumulate unchecked across column blocks (red). In REAL-Q, a dynamic Block-GD step is applied immediately after each block, actively correcting accumulated structural errors and keeping the end-to-end reconstruction loss tightly bounded (green).}
    \label{fig:blockgd}
\end{figure}

\subsection{Objective}
\label{sec:fisher_loss}

A challenge in moving beyond static analytical solvers is defining a tractable, high-fidelity loss function to drive dynamic optimization. Using the exact end-to-end KL divergence (Eq.~\ref{eq:kl_loss}) as an explicit objective would require a full forward pass through all subsequent layers and the LM head at every block step---a computationally prohibitive requirement.

To break this computational deadlock, we construct a high-fidelity surrogate: the \emph{Aggregated Fisher MSE loss}. Consider the KL divergence $\cL_{\text{KL}}$ as a function of the per-token transformer block output $\by_t$. By performing a second-order Taylor expansion around the full-precision output $\by_t^*$, and noting that the first-order gradient identically vanishes since $\by_t^*$ minimizes the KL divergence ($\cL_{\text{KL}}(\by_t^*) = 0$), the expected loss over tokens satisfies:
\begin{equation}
    \EE_t\bigl[\cL_{\text{KL}}(\by_t)\bigr] \approx \frac{1}{2} \EE_t\bigl[\Delta \by_t^\top \bH_t \Delta \by_t\bigr],
    \label{eq:taylor_expansion}
\end{equation}
where $\Delta \by_t = \by_t - \by_t^*$ is the per-token output perturbation and $\bH_t = \nabla^2_{\by_t} \cL_{\text{KL}}\big|_{\by_t^*}$ is the per-token Hessian. The standard Fisher--Hessian identity gives $\EE_t[\bH_t] = \EE_t[\bg_t \bg_t^\top]$ in expectation under the model distribution, where $\bg_t = \nabla_{\by_t} \cL_{\text{NLL}}$ (noting the shift from $\cL_{\text{KL}}$, as Fisher information is formally defined via the NLL score function using target labels sampled from the full-precision model's output distribution $p$). In contrast to the structural approximations used by prior analytical solvers, we formulate an objective that operates at the \emph{transformer block} level and preserves the cross-channel coupling structure.

\paragraph{Aggregated Fisher loss (ours).} We bypass the structural approximations of prior works by moving the expectation \emph{inside} the quadratic form. We define the aggregated Fisher matrix:
\begin{equation}
    \bF = \EE_t\bigl[\bg_t \bg_t^\top\bigr] \approx \frac{1}{T}\sum_{t=1}^{T} \bg_t \bg_t^\top \in \RR^{d \times d},
    \label{eq:fisher_full}
\end{equation}
where $d$ is the transformer block output dimension. Our surrogate loss, which acts as the foundational objective for our subsequent Block-GD updates (Section~\ref{sec:block_gd}), is formulated as:
\begin{equation}
    \cL_{\text{Fisher}} = \frac{1}{2} \EE_t\bigl[\Delta \by_t^\top \bF \Delta \by_t\bigr] = \frac{1}{2} \cdot \frac{1}{T}\sum_{t=1}^{T} \Delta \by_t^\top \bF \Delta \by_t.
    \label{eq:fisher_loss}
\end{equation}
Here, $\bF$ is the \emph{full} Fisher matrix (retaining all cross-channel covariances), aggregated once per transformer block over the calibration set. It requires zero per-token storage and removes the need for any heuristic grouping hyperparameters. Details regarding the pre-computation of $\bF$ and the special handling of the final transformer block are provided in Appendix~\ref{app:exp_details_full}.

\begin{proposition}[Decoupled Fisher Aggregation]
\label{prop:fisher_decoupled}
The exact expected second-order penalty is $\EE_t[\Delta\by_t^\top \bH_t \Delta\by_t]$. By applying a mean-field approximation to decouple the token-wise Hessian from the output perturbation, we obtain $\EE_t[\Delta\by_t^\top\, \EE_t[\bH_t]\, \Delta\by_t]$. Combined with the Fisher--Hessian identity $\EE_t[\bH_t] = \EE_t[\bg_t\bg_t^\top] \approx \bF$, Eq.~\ref{eq:fisher_loss} serves as a tractable surrogate. The approximation error strictly amounts to the element-wise covariance $\sum_{i,j} \text{Cov}\bigl((\bH_t)_{ij}, (\Delta\by_t \Delta\by_t^\top)_{ij}\bigr)$ (see Appendix~\ref{app:fisher_proof}).
\end{proposition}

We adopt this decoupling as a deliberate trade-off: recognizing that cross-channel covariance affects LLM behavior more strongly than token-wise correlations, we trade the latter to bypass intractable per-token Hessians. This allows materializing the full aggregated Fisher $\bF$ without memory explosion.

\subsection{Dynamic Block-wise Gradient Descent}
\label{sec:block_gd}

In REAL-Q, the saliency-weighted analytical update serves as a foundational step. After analytically processing column block $b$, we evaluate our surrogate $\cL_{\text{Fisher}}(\bW^{(b)})$ via a forward pass through the current transformer block using a calibration mini-batch. We explicitly correct the static solver's residual errors by computing the exact gradient with respect to the remaining unquantized weights, updating these trailing columns via a single Adam \citep{kingma2014adam} step (whose states are tracked per-parameter, as the pool of active unquantized weights continuously shrinks):
\begin{equation}
    \bW_{:, b_{\text{end}}:n} \leftarrow \bW_{:, b_{\text{end}}:n} - \eta \cdot \text{Adam}\big(\nabla_{\bW_{:, b_{\text{end}}:n}} \cL_{\text{Fisher}}(\bW^{(b)})\big),
    \label{eq:block_gd_update}
\end{equation}
where $\eta$ is the learning rate. Already-quantized weights remain locked. By relying on this first-order correction, REAL-Q limits cross-block error accumulation and avoids the prohibitive cost of refreshing the analytical Hessian during the quantization loop.

\subsection{Loss Sliding Window}
\label{sec:slide_window}

Sequential quantization typically suffers from discontinuous objective landscapes when transitioning between transformer blocks. To resolve this, we introduce a \emph{loss sliding window} (Figure~\ref{fig:slidewindow}) that smoothly interpolates exact Fisher objectives. When quantizing block $\ell$ at cumulative column-block step $b_{\text{global}}$ (out of $B_{\text{total}}$, where $B_{\text{total}}$ denotes the total number of column-block steps within transformer block $\ell$, i.e., the sum of column blocks across all linear modules in the block; the per-linear-module column-block index $b$ in Eq.~\ref{eq:block_gd_update} is local to its module, while $b_{\text{global}}$ accumulates across all modules in the block), the surrogate is dynamically blended:
\begin{equation}
    \cL_{\text{slide}}^{(b_{\text{global}})} = \alpha^{(b_{\text{global}})} \cdot \cL_{\text{Fisher}}^{(\ell)} + \bigl(1 - \alpha^{(b_{\text{global}})}\bigr) \cdot \cL_{\text{Fisher}}^{(\ell+1)},
    \label{eq:slide_window}
\end{equation}
where $\alpha^{(b_{\text{global}})} = 1 - \frac{b_{\text{global}}-1}{B_{\text{total}} - 1}$ (assuming $B_{\text{total}} > 1$). Mechanically, evaluating the gradient of this blended loss requires backpropagating from the output of block $\ell+1$ into block $\ell$. While this expands the dynamic correction graph to two transformer blocks, it avoids the overhead of full end-to-end unrolling. This two-block scope encourages the current layer to anticipate and reduce error propagation into its successor, preventing abrupt jumps in the loss landscape between adjacent blocks and thereby stabilizing the sequential quantization process.

\section{Theoretical Analysis}
\label{sec:theory}

We present two theoretical results justifying REAL-Q: formalizing the inherent error accumulation of static Hessian solvers, and deriving a sufficient cosine condition under which Block-GD descends on the true global loss. Full proofs are deferred to Appendix~\ref{app:error_accu} and~\ref{app:blockgd_bounds}.

\paragraph{Error Accumulation in Static Solvers.}
Globally-guided PTQ methods rely on a static approximation of the end-to-end Hessian ($\bH$). However, as column quantization progresses, the true downstream-aware Hessian shifts. Letting $\Delta \bH^{(j)} = \bH_{\mathrm{ideal}}^{(j)} - \bH$ denote this gap, the static solver incurs a cumulative penalty. Assuming a linear structural drift, we prove in Appendix~\ref{app:error_accu} that this expected uncompensated error scales as:
\begin{equation}
    \EE[\mathcal{E}] = \Omega(n \cdot \bar{\epsilon}^2 \cdot \|\bH\|_2),
\end{equation}
where $n$ is the column dimension and $\bar{\epsilon}^2$ is the expected squared residual. Under the stated assumption, this linear error growth with respect to module width $n$ indicates that static analytical solvers cannot prevent compounding cross-column errors.

\paragraph{Descent Condition Analysis under First-Order Updates.}
To overcome this, REAL-Q dynamically optimizes the surrogate $\cL_{\text{Fisher}}$. Assuming the true end-to-end objective $J(\bW)$ is $\beta$-smooth, the true loss after a gradient step $\bW^{(+)} = \bW - \eta \bg_{\text{surr}}$ satisfies the descent-lemma upper bound
\begin{equation}
    J(\bW^{(+)}) \leq J(\bW) - \eta \|\nabla J(\bW)\| \|\bg_{\text{surr}}\| \cos \theta + \frac{\eta^2 \beta}{2} \|\bg_{\text{surr}}\|^2,
\end{equation}
where $\cos \theta$ is the cosine similarity between the surrogate gradient $\bg_{\text{surr}}$ and the true gradient $\nabla J(\bW)$. Thus, the true global loss strictly decreases ($J(\bW^{(+)}) < J(\bW)$) if:
\begin{equation}
    \cos \theta > \frac{\eta \beta}{2} \cdot \frac{\|\bg_{\text{surr}}\|}{\|\nabla J(\bW)\|}.
    \label{eq:exact_cosine_bound}
\end{equation}
This establishes that optimization fails under poor gradient approximations ($\cos \theta \sim 0$). Conversely, REAL-Q maintains $\cos \theta$ well above this threshold (Figure~\ref{fig:cosine_body}). The above argument is stated for SGD; for Adam, the preconditioned step direction is no longer aligned with $\bg_{\text{surr}}$, so the inequality does not directly transfer, but we observe consistent empirical descent under the learning rates used in our experiments (Section~\ref{sec:experiments}).

\section{Experiments}
\label{sec:experiments}

\subsection{Experimental Setup}
\label{sec:exp_setup}

\paragraph{Setup.} We evaluate REAL-Q on LLaMA-3.1-8B / LLaMA-3.1-70B \citep{grattafiori2024llama} and Qwen3-0.6B / 1.7B / 4B / 8B / 32B \citep{yang2025qwen3}, against RTN, GPTQ \citep{frantar2023gptq}, GPTAQ \citep{li2025gptaq}, and GuidedQuant \citep{kim2025guidedquant}; all methods share QuaRot \citep{ashkboos2024quarot} rotation. We report WikiText-2 \citep{merity2017wikitext} KL divergence, perplexity, and zero-shot accuracy on ten tasks. Quantization is symmetric throughout; W4A16 uses \emph{per-row} weights (no grouping); group size 128 is used only for low-bit (W3A16, W2A16) and weight-activation (W$x$A4KV4) settings. Calibration: 2048 WikiText-2 samples for W4A16 and W4A4KV4 (256 for LLaMA-3.1-70B due to cost; 256 for low-bit / W2A4KV4 / W3A4KV4). We use Adam with a mini-batch size of 32 and a reverse-cosine layer-wise LR schedule. Full setup and learning rates are detailed in Appendices~\ref{app:exp_details_full} and~\ref{app:realq_lr}.

\paragraph{Why KL is our primary fidelity metric.} \textbf{All Qwen3 checkpoints used in our evaluation are post-trained models, for which the bf16 reference is generally not at the extremum of perplexity or zero-shot task accuracy} (post-training optimizes neither metric; cf.\ Appendix~\ref{app:exp_details_full}). Consequently, any quantization-induced fluctuation in PPL or downstream accuracy---in either direction---does not by itself indicate quantization quality; only the KL divergence between the quantized and full-precision output distributions directly measures fidelity to the full-precision model's behavior. \textbf{Furthermore, the calibration set ($2048 \times 2048 \approx 4\text{M}$ tokens) is a small fraction of the full WikiText-2 corpus, and KL/PPL are evaluated on the held-out test split}, so the consistent KL reduction across every evaluated model reflects generalization to unseen text rather than overfitting to the calibration sample.

\subsection{Main Results: W4A16}
\label{sec:exp_w4a16}

Table~\ref{tab:w4a16_body} reports W4A16 results (per-row weight quantization) for REAL-Q and four baselines on four representative models---one LLaMA-3.1 base model and three Qwen3 sizes. REAL-Q achieves the lowest KL divergence on every model. The full seven-model table (additionally covering LLaMA-3.1-70B, Qwen3-0.6B / 8B) is in Appendix~\ref{app:exp_full}.

\begin{table}[h]
    \vspace{-0.3cm}
    \caption{W4A16 (per-row weight quantization) on four representative models. KL ($\times 10^{-2}$) and PPL are evaluated on WikiText-2. Downstream columns report zero-shot accuracy (\%) on ten tasks and their average. \textbf{KL is the primary fidelity metric}; for post-trained Qwen3 models the bf16 reference is not at the PPL/accuracy optimum, so PPL or downstream-accuracy fluctuations relative to bf16 do not by themselves indicate quality (see \S\ref{sec:exp_setup} and Appendix~\ref{app:exp_details_full}).}
    \label{tab:w4a16_body}
    \centering
    \scriptsize
    \setlength{\tabcolsep}{2pt}
    \begin{tabular}{ll*{13}{c}}
        \toprule
        Model & Method & KL & PPL & ARC-C & ARC-E & BoolQ & C-Eval & Hella. & LAMB. & OBQA & PIQA & SIQA & Wino. & Avg. \\
        \midrule
        \multirow{6}{*}{LLaMA-3.1-8B}
            & bf16 & --- & 6.25 & 53.41 & 81.31 & 82.17 & 48.66 & 78.90 & 75.33 & 44.80 & 81.18 & 50.15 & 73.80 & 66.97 \\
            & RTN & 16.1 & 7.64 & 49.23 & 76.81 & 79.24 & 38.86 & 75.08 & 70.33 & 43.60 & 78.07 & 49.85 & 71.67 & 63.27 \\
            & GPTQ & 4.95 & 6.60 & 53.75 & 80.01 & 81.53 & 46.51 & \textbf{78.20} & \textbf{75.02} & 44.20 & 80.58 & 49.59 & 72.45 & 66.18 \\
            & GPTAQ & 3.98 & 6.53 & 53.16 & \textbf{80.60} & 79.82 & 45.99 & 78.11 & 74.46 & 43.80 & 80.20 & 50.41 & \textbf{73.09} & 65.96 \\
            & GuidedQuant & 3.67 & 6.50 & 53.07 & 79.97 & \textbf{81.74} & \textbf{46.95} & 77.96 & 74.35 & 44.20 & 80.47 & 49.69 & 72.85 & 66.12 \\
            \rowcolor{gray!15} & REAL-Q (ours) & \textbf{3.36} & \textbf{6.48} & \textbf{53.84} & 79.97 & 81.07 & 46.36 & 78.08 & 74.91 & \textbf{46.20} & \textbf{80.96} & \textbf{50.61} & 72.85 & \textbf{66.48} \\
        \midrule
        \multirow{6}{*}{Qwen3-1.7B}
            & bf16 & --- & 16.73 & 42.75 & 69.91 & 77.95 & 58.62 & 60.37 & 50.44 & 36.80 & 72.47 & 44.22 & 60.93 & 57.45 \\
            & RTN & 120 & 75.61 & 30.72 & 37.67 & 64.80 & 42.20 & 49.67 & 24.49 & 29.20 & 63.00 & 41.56 & 54.06 & 43.74 \\
            & GPTQ & 21.3 & 21.50 & 34.13 & 56.36 & 76.91 & 52.23 & 56.94 & 42.60 & 34.40 & 67.74 & 43.30 & 57.46 & 52.21 \\
            & GPTAQ & 13.2 & 18.59 & 37.71 & 60.35 & 75.54 & 51.11 & 57.35 & 46.09 & 34.00 & 70.35 & 42.73 & 59.67 & 53.49 \\
            & GuidedQuant & 11.9 & 17.72 & \textbf{39.68} & \textbf{65.87} & 76.39 & 48.51 & 58.23 & 41.70 & \textbf{37.60} & \textbf{70.40} & \textbf{43.96} & \textbf{60.06} & 54.24 \\
            \rowcolor{gray!15}& REAL-Q (ours) & \textbf{6.07} & \textbf{16.54} & 38.40 & 58.67 & \textbf{77.98} & \textbf{55.42} & \textbf{58.67} & \textbf{48.22} & 36.20 & 70.29 & 42.84 & 59.75 & \textbf{54.64} \\
        \midrule
        \multirow{6}{*}{Qwen3-4B}
            & bf16 & --- & 13.66 & 54.01 & 78.41 & 85.20 & 70.43 & 68.40 & 59.29 & 40.40 & 75.03 & 51.28 & 66.14 & 64.86 \\
            & RTN & 31.1 & 17.11 & 46.76 & 69.82 & 80.46 & 62.85 & 64.80 & 50.42 & 38.40 & 73.29 & 48.41 & 62.67 & 59.79 \\
            & GPTQ & 13.5 & 13.58 & \textbf{50.26} & 74.20 & 83.49 & 61.89 & 66.67 & 59.07 & \textbf{40.20} & 74.70 & 48.46 & 63.93 & 62.29 \\
            & GPTAQ & 11.2 & 14.59 & 48.89 & 73.61 & \textbf{83.58} & 64.12 & 65.87 & 58.45 & 39.60 & 74.92 & \textbf{49.33} & 62.90 & 62.13 \\
            & GuidedQuant & 9.50 & 14.50 & 49.23 & 74.16 & 82.75 & 64.49 & 66.55 & 57.31 & 38.40 & 74.43 & 48.62 & 62.35 & 61.83 \\
            \rowcolor{gray!15}& REAL-Q (ours) & \textbf{5.44} & \textbf{13.44} & 49.74 & \textbf{74.62} & 83.46 & \textbf{67.38} & \textbf{67.06} & \textbf{60.12} & 39.40 & \textbf{75.90} & 46.32 & \textbf{65.11} & \textbf{62.91} \\
        \midrule
        \multirow{6}{*}{Qwen3-32B}
            & bf16 & --- & 7.61 & 60.75 & 83.38 & 86.82 & 85.96 & 82.58 & 67.24 & 46.20 & 81.94 & 54.04 & 72.93 & 72.18 \\
            & RTN & 21.8 & 8.57 & 58.02 & 77.31 & 86.67 & 82.54 & 80.30 & 64.45 & 47.40 & 79.54 & 50.82 & 71.90 & 69.90 \\
            & GPTQ & 4.98 & 7.74 & 58.87 & 81.61 & 86.39 & \textbf{85.14} & \textbf{82.32} & 67.05 & 46.20 & 81.56 & 53.12 & 72.14 & 71.44 \\
            & GPTAQ & 6.62 & 7.84 & 60.92 & 83.38 & 86.45 & 84.84 & 81.61 & \textbf{67.88} & 45.00 & 81.07 & 53.53 & \textbf{73.56} & 71.82 \\
            & GuidedQuant & 4.66 & 7.74 & \textbf{62.29} & \textbf{83.42} & 83.88 & 84.47 & 82.15 & 67.71 & \textbf{48.20} & \textbf{81.61} & \textbf{54.20} & 72.30 & \textbf{72.02} \\
            \rowcolor{gray!15}& REAL-Q (ours) & \textbf{4.24} & \textbf{7.53} & 60.84 & 83.29 & \textbf{87.43} & 84.92 & 82.19 & 67.34 & 45.40 & 81.39 & 51.84 & 72.38 & 71.70 \\
        \bottomrule
    \end{tabular}
    \vspace{-0.3cm}
\end{table}

\subsection{Low-bit Weight-only Quantization}
\label{sec:exp_low_bit}

Table~\ref{tab:low_bit} compares baselines and REAL-Q on Qwen3-8B at W3A16 and W2A16. REAL-Q still beats every baseline including GuidedQuant.

\begin{table}[H]
    \vspace{-0.3cm}
    \caption{Low-bit weight-only quantization on Qwen3-8B. Group size 128, calibration sequence length 2048, 256 calibration samples. KL ($\times 10^{-2}$) and PPL are evaluated on WikiText-2. Downstream columns report zero-shot accuracy (\%) on ten tasks and their average.}
    \label{tab:low_bit}
    \centering
    \scriptsize
    \setlength{\tabcolsep}{1.8pt}
    \begin{tabular}{ll*{13}{c}}
        \toprule
        Model & Method & KL & PPL & ARC-C & ARC-E & BoolQ & C-Eval & Hella. & LAMB. & OBQA & PIQA & SIQA & Wino. & Avg. \\
        \midrule
        \multirow{5}{*}{W3A16}
            & RTN & 74.1 & 19.58 & 38.14 & 62.16 & 73.00 & 51.11 & 56.30 & 35.78 & 34.60 & 70.62 & 42.27 & 57.77 & 52.17 \\
            & GPTQ & 13.0 & 10.70 & 51.96 & 77.44 & 84.53 & 70.06 & 70.13 & \textbf{63.67} & \textbf{41.00} & 75.95 & \textbf{49.95} & 67.80 & 65.25 \\
            & GPTAQ & 12.2 & 10.71 & 52.13 & 77.57 & 85.44 & 69.17 & 70.23 & 62.10 & 39.40 & 75.24 & 48.31 & 67.48 & 64.71 \\
            & GuidedQuant & 11.6 & 10.38 & 50.68 & 78.07 & 84.50 & 70.36 & 70.75 & 61.50 & 40.20 & 76.28 & 48.93 & 66.85 & 64.81 \\
            \rowcolor{gray!15}& REAL-Q (ours) & \textbf{10.8} & \textbf{10.29} & \textbf{52.56} & \textbf{79.12} & \textbf{85.78} & \textbf{70.58} & \textbf{70.79} & 61.19 & 40.00 & \textbf{76.66} & 49.49 & \textbf{68.43} & \textbf{65.50} \\
        \midrule
        \multirow{5}{*}{W2A16}
            & RTN & 541 & $1.09\!\times\!10^8$ & 25.85 & 24.24 & 39.27 & 26.15 & 26.66 & 0.00 & 27.40 & 51.47 & 33.62 & 50.51 & 30.52 \\
            & GPTQ & 62.6 & 17.21 & 27.90 & 43.39 & 52.26 & 27.19 & 47.03 & 27.03 & 30.80 & 63.06 & 37.56 & 54.38 & 41.06 \\
            & GPTAQ & 60.6 & 16.60 & 29.86 & 45.20 & 67.80 & 25.63 & 49.37 & 33.75 & 31.60 & 64.53 & 38.84 & 53.99 & 44.06 \\
            & GuidedQuant & 58.8 & 15.91 & 26.79 & 43.56 & 60.76 & \textbf{27.49} & 46.37 & 24.32 & 29.40 & 62.24 & 37.26 & 52.41 & 41.06 \\
            \rowcolor{gray!15}& REAL-Q (ours) & \textbf{49.9} & \textbf{14.94} & \textbf{34.56} & \textbf{61.57} & \textbf{72.05} & 26.52 & \textbf{54.32} & \textbf{38.35} & \textbf{36.20} & \textbf{68.66} & \textbf{41.76} & \textbf{58.80} & \textbf{49.28} \\
        \bottomrule
    \end{tabular}
    \vspace{-0.1cm}
\end{table}

\subsection{Weight-Activation Quantization}
\label{sec:exp_wa}

We evaluate REAL-Q in the weight-activation quantization setting on Qwen3-8B with W4A4KV4, W3A4KV4 and W2A4KV4 configurations. Activations and KV cache are quantized per-token with clipping ratios $a_{\text{clip}}=k_{\text{clip}}=v_{\text{clip}}=0.9$; weights use group size 128. As described in Appendix~\ref{app:exp_details_full}, the reverse-cosine learning-rate schedule is disabled specifically in activation-aware settings, ensuring that early-block errors are corrected by gradient descent rather than amplified.

\paragraph{Two activation-quantization variants.} For each method we report two activation-quantization variants. \textbf{(a) Activation-unaware:} weights are quantized using the full-precision activations as input; the activations are quantized only \emph{after} all weights are quantized. \textbf{(b) Activation-aware:} the quantization of weights is performed against already-quantized activations, so each weight column sees the same input distribution it will face at inference. GPTAQ and GuidedQuant both natively support activation-aware quantization, and we therefore enable that variant for those baselines.

\begin{table}[H]
    \vspace{-0.3cm}
    \caption{Weight-activation quantization on Qwen3-8B (W$x$A4KV4). KL ($\times 10^{-2}$) and PPL are evaluated on WikiText-2. Downstream columns report zero-shot accuracy (\%) on ten tasks and their average. For REAL-Q we report both the activation-unaware (a-unaware) and the activation-aware (a-aware) variants. GPTQ / GPTAQ / GuidedQuant baselines use their activation-aware variant (where supported).}
    \label{tab:w_a_quant}
    \centering
    \scriptsize
    \setlength{\tabcolsep}{1.6pt}
    \begin{tabular}{ll*{13}{c}}
        \toprule
        Model & Method & KL & PPL & ARC-C & ARC-E & BoolQ & C-Eval & Hella. & LAMB. & OBQA & PIQA & SIQA & Wino. & Avg. \\
        \midrule
        \multirow{6}{*}{W4A4KV4}
            & RTN & 40.8 & 13.09 & 38.40 & 61.99 & 76.97 & 61.89 & 64.17 & 52.44 & 37.80 & 71.11 & 44.78 & 59.83 & 56.94 \\
            & GPTQ & 27.0 & 11.50 & 47.95 & 71.00 & 82.08 & 64.64 & 67.67 & 54.41 & 39.20 & 71.98 & 47.54 & 61.40 & 60.79 \\
            & GPTAQ & 27.1 & 12.06 & 45.48 & 69.11 & 81.83 & 64.49 & 67.53 & 53.00 & 39.40 & 73.39 & \textbf{48.16} & 61.72 & 60.41 \\
            & GuidedQuant & 27.0 & 11.54 & 47.35 & 72.81 & 81.10 & \textbf{67.38} & 67.25 & 55.40 & 37.80 & 73.07 & 47.54 & 61.33 & 61.10 \\
            & REAL-Q (a-unaware) & 26.2 & 11.42 & 49.57 & 72.14 & 80.98 & 63.60 & 67.66 & 54.05 & 37.60 & 74.54 & 45.70 & \textbf{63.93} & 60.98 \\
            \rowcolor{gray!15}& REAL-Q (a-aware) & \textbf{22.8} & \textbf{11.12} & \textbf{51.02} & \textbf{75.63} & \textbf{83.67} & 66.12 & \textbf{69.05} & \textbf{58.86} & \textbf{39.60} & \textbf{75.52} & 46.88 & 63.61 & \textbf{63.00} \\
        \midrule
        \multirow{6}{*}{W3A4KV4}
            & RTN & 121 & 35.16 & 29.44 & 48.99 & 56.36 & 33.73 & 45.93 & 18.69 & 29.60 & 62.57 & 38.23 & 52.64 & 41.62 \\
            & GPTQ & 36.7 & 12.69 & 43.43 & 67.59 & 74.01 & 56.17 & 62.98 & 51.27 & 36.60 & 70.62 & 45.85 & 58.80 & 56.73 \\
            & GPTAQ & 36.4 & 13.10 & 42.49 & 67.21 & 79.54 & 53.12 & 63.65 & 48.50 & 37.60 & \textbf{72.58} & 44.01 & 60.38 & 56.91 \\
            & GuidedQuant & 35.2 & 12.40 & 44.37 & 71.51 & 80.80 & 56.02 & 63.25 & 48.81 & 37.20 & 71.60 & 44.88 & 58.96 & 57.74 \\
            & REAL-Q (a-unaware) & 32.1 & \textbf{12.00} & 44.37 & \textbf{72.60} & 78.65 & 57.28 & 64.50 & 50.05 & 37.60 & 71.60 & 45.04 & 61.25 & 58.29 \\
            \rowcolor{gray!15}& REAL-Q (a-aware) & \textbf{30.5} & 12.01 & \textbf{45.73} & 71.13 & \textbf{81.38} & \textbf{58.25} & \textbf{65.15} & \textbf{55.11} & \textbf{39.00} & 72.36 & \textbf{45.39} & \textbf{62.35} & \textbf{59.59} \\
        \midrule
        \multirow{6}{*}{W2A4KV4}
            & RTN & 549 & $1.45\!\times\!10^8$ & 26.28 & 24.20 & 38.35 & 26.23 & 26.49 & 0.00 & 28.40 & 52.18 & 33.98 & 50.59 & 30.67 \\
            & GPTQ & 133 & 48.73 & 24.06 & 35.77 & 40.73 & 22.96 & 34.27 & 6.35 & 25.60 & 56.80 & 35.11 & 51.54 & 33.32 \\
            & GPTAQ & 108 & 28.13 & 24.66 & 34.05 & 53.06 & 25.04 & 37.83 & 16.24 & 26.80 & 57.83 & 34.44 & \textbf{55.09} & 36.50 \\
            & GuidedQuant & 106 & 30.75 & 23.81 & 37.25 & 47.86 & 23.33 & 36.90 & 7.53 & 23.60 & 56.37 & 35.01 & 52.88 & 34.45 \\
            & REAL-Q (a-unaware) & 127 & 42.56 & 27.56 & 43.52 & 48.90 & 23.25 & 41.62 & 12.71 & 25.80 & 60.50 & 36.28 & 52.01 & 37.21 \\
            \rowcolor{gray!15}& REAL-Q (a-aware) & \textbf{76.7} & \textbf{19.68} & \textbf{29.52} & \textbf{47.22} & \textbf{65.50} & \textbf{27.04} & \textbf{47.62} & \textbf{30.53} & \textbf{31.40} & \textbf{64.36} & \textbf{38.54} & 51.14 & \textbf{43.29} \\
        \bottomrule
    \end{tabular}
    \vspace{-0.5cm}
\end{table}

\subsection{Ablation on Qwen3-4B}
\label{sec:exp_ablation}

We ablate four REAL-Q design choices on Qwen3-4B (W4A16, per-row weight quantization, scheduled final learning rate $2\!\times\!10^{-4}$, final-block LR $10^{-5}$): the loss \textbf{sliding window} (SW, on/off); the output-perturbation clipping ratio \textbf{$a_{\text{loss\_clip}}$}; the column \textbf{block size} $B$ (smaller $B$ $\Rightarrow$ more frequent gradient corrections); and the number of \textbf{backward samples} per gradient step (controls the variance of $\bg^{(b)}$). The base configuration is SW on, $a_{\text{loss\_clip}}{=}0.95$, $B{=}128$, 32 backward samples per step.

\begin{table}[H]
    \vspace{-0.4cm}
    \caption{Ablation on Qwen3-4B (W4A16, per-row weight quantization). KL ($\times 10^{-2}$) and PPL are evaluated on WikiText-2; Avg.\ is the average zero-shot accuracy across the ten downstream tasks. Each row changes one knob from the base; remaining knobs are at the base values.}
    \label{tab:ablation_4b}
    \centering
    \small
    \setlength{\tabcolsep}{4pt}
    \begin{tabular}{lccccccc}
        \toprule
        Configuration & SW & $a_{\text{loss\_clip}}$ & $B$ & samples & KL & PPL & Avg. \\
        \midrule
        \rowcolor{gray!15}Base             & on  & 0.95 & 128 & 32 & \textbf{5.44} & 13.44 & 62.91 \\
        $-$ slide window            & off & 0.95 & 128 & 32 & 5.64 & 13.32 & 62.95 \\
        $-$ activation clip         & on  & --   & 128 & 32 & 5.74 & \textbf{13.28} & 62.81 \\
        Larger column block size    & on  & 0.95 & 512 & 32 & 6.57 & 13.76 & 62.62 \\
        Fewer backward samples      & on  & 0.95 & 128 & 8  & 5.72 & 13.45 & \textbf{63.53} \\
        \bottomrule
    \end{tabular}
    \vspace{-0.4cm}
\end{table}

Notably, the column block size $B$ exhibits the largest single-knob impact, supporting the role of fine-grained update granularity (smaller $B$) in mitigating error accumulation. We use KL as the primary metric throughout (\S\ref{sec:exp_setup}).

\subsection{Loss Approximation Quality}
\label{sec:cosine_body}

\begin{figure}[h]
    \centering
    \includegraphics[width=\linewidth]{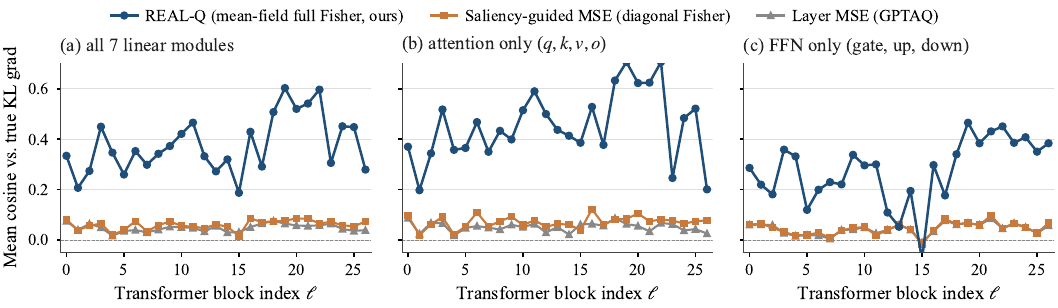}
    \caption{Mean cosine similarity between each surrogate's gradient and the true end-to-end KL gradient on Qwen3-0.6B (W4A16). REAL-Q's full Fisher (blue) remains consistently aligned with the true gradient, whereas the saliency-guided MSE (GuidedQuant, orange) and module MSE (GPTQ / GPTAQ, grey) sit near zero alignment.}
    \label{fig:cosine_body}
     \vspace{-0.4cm}
\end{figure}

To empirically assess the quality of our aggregated Fisher MSE as a surrogate, we evaluate the geometric alignment between the surrogate gradients and the true end-to-end KL gradient. Figure~\ref{fig:cosine_body} plots this mean cosine similarity on Qwen3-0.6B (W4A16). REAL-Q maintains high alignment with the true global trajectory. By contrast, the layer-wise MSE (GPTQ/GPTAQ) and saliency-guided MSE (GuidedQuant) hover near zero, consistent with their limited ability to capture the true descent direction. Full experimental setup is detailed in Appendix~\ref{app:cosine_exp}.

\paragraph{Additional results.} Full W4A16 sweeps, calibration-seed stability on Qwen3-0.6B, and detailed wall-clock and GPU memory profiling are deferred to Appendices~\ref{app:exp_full},~\ref{app:seed_variance},~\ref{app:gpu_hours}, and~\ref{app:memory_usage}.

\section{Discussion, Limitations, and Future Work}
\label{sec:discussion}

REAL-Q addresses the precision--tractability trade-off in modern PTQ by replacing static analytical solvers with a dynamic, coarse-to-fine optimization hierarchy. By pairing a full Aggregated Fisher MSE with Block-GD, our method reduces \emph{information misalignment} and preserves cross-channel coupling. Furthermore, our theoretical analysis shows that, under a linear structural-drift assumption, static solvers incur $\Omega(n)$ compounding errors, and gives the cosine condition that, when satisfied, guarantees descent of our dynamic first-order updates on the true global loss. Shifting to this active gradient-based paradigm introduces additional computational and memory overhead during calibration; we expect these overheads can be amortized through future system-level optimizations, such as custom memory-efficient backward kernels and intermediate activation offloading. An extended discussion of limitations and future work is provided in Appendix~\ref{app:limitations}.

\bibliography{references}
\bibliographystyle{plainnat}


\newpage
\appendix

\section{REAL-Q Algorithm Pseudocode and Sliding-Window Illustration}
\label{app:algorithm}

\begin{algorithm}[htbp]
\caption{REAL-Q Quantization}
\label{alg:gptqplus}
\begin{algorithmic}[1]
\REQUIRE Pre-trained model with weights $\{\bW^{(\ell)}\}$, calibration set $\cD$, block size $B$, learning rate $\eta$
\STATE \textbf{Stage 0:} Run end-to-end backward on $\cD$ to compute aggregated Fisher matrix $\mathbf{F}^{(\ell)}$ and saliency $\mathbf{s}^{(\ell)}$ for all layers $\ell$
\FOR{layer $\ell = 1, \ldots, L$}
    \FOR{each linear module in layer $\ell$}
        \STATE Compute saliency-weighted Hessian $\bH_g$ and its inverse $(\bH_g)^{-1}$ for all row-groups $g$
        \FOR{block $b = 1, \ldots, \lceil n/B \rceil$}
            \STATE \textit{// Standard GPTQ within block (executed parallelly across groups)}
            \FOR{column $j$ in block $b$}
                \STATE $\hat{\mathbf{w}}_{g, j} \leftarrow \text{quant}(\bW_{g, j})$; \quad $\boldsymbol{\delta}_{g, j} \leftarrow \bW_{g, j} - \hat{\mathbf{w}}_{g, j} \quad (\forall g)$
                \STATE Update remaining columns in block via Eq.~\ref{eq:gptq_update} \quad $(\forall g)$
            \ENDFOR
            \STATE \textit{// Second-order compensation for trailing columns}
            \STATE $\bW_{g, b_{\text{end}}:n} \leftarrow \bW_{g, b_{\text{end}}:n} - \mathbf{E}_{g, b} \, (\bH_g^{-1})_{b, b_{\text{end}}:n} \quad (\forall g)$
            \STATE \textit{// Block-wise gradient descent (jointly updates the full unquantized matrix)}
            \STATE \textit{// Note: $b_{\text{global}}$ is the cumulative column-block step within the current transformer block, summed across all linear modules in the block (matches the index $b$ in Eq.~\ref{eq:slide_window}).}
            \IF{$\ell = L$}
                \STATE $\cL^{(b)} \leftarrow$ true KL divergence $\cL_{\text{KL}}$ against the LM head
            \ELSE
                \STATE $\cL^{(b)} \leftarrow$ sliding-window loss $\cL_{\text{slide}}^{(b_{\text{global}})}$ (Eq.~\ref{eq:slide_window}) using Fisher MSE
            \ENDIF
            \STATE $\bg^{(b)} \leftarrow \nabla_{\bW_{:, b_{\text{end}}:n}} \cL^{(b)}$
            \STATE $\bW_{:, b_{\text{end}}:n} \leftarrow \bW_{:, b_{\text{end}}:n} - \eta \cdot \text{Adam}(\bg^{(b)})$
        \ENDFOR
    \ENDFOR
\ENDFOR
\end{algorithmic}
\end{algorithm}

\begin{figure}[htbp]
    \centering
    \includegraphics[width=\linewidth]{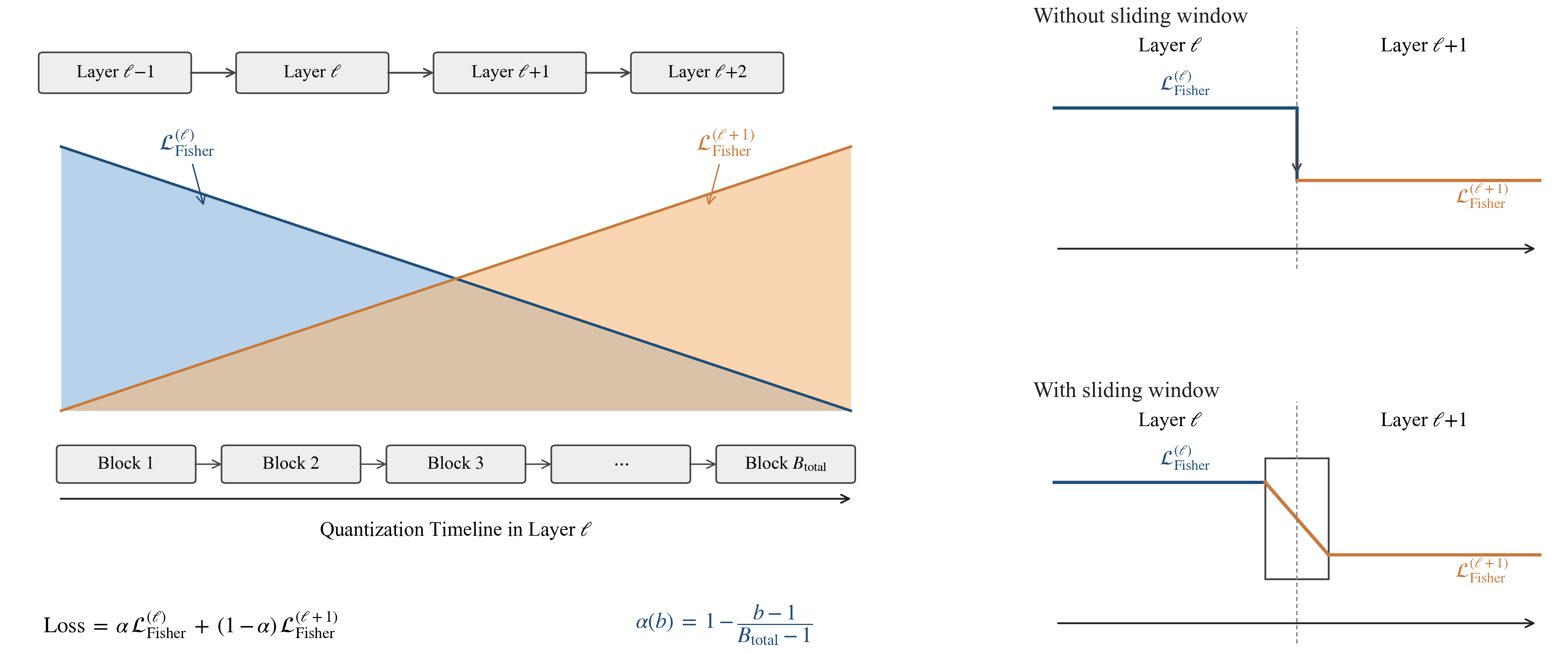}
    \caption{Illustration of the loss sliding window mechanism (referenced from Section~\ref{sec:slide_window}). When quantizing transformer block $\ell$, the weight of $\cL_{\text{Fisher}}^{(\ell)}$ (blue) linearly decreases while $\cL_{\text{Fisher}}^{(\ell+1)}$ (orange) linearly increases across the column blocks of the linear modules in transformer block $\ell$. \textbf{Right:} without the sliding window, the objective exhibits a sharp discontinuity at the transformer-block boundary; with it, the transition is a smooth linear interpolation.}
    \label{fig:slidewindow}
\end{figure}

\section{Extended Related Work}
\label{app:related_extra}

\paragraph{Weight-only PTQ methods.} A complementary line of work targets aggressive low-bit weight-only quantization, where activations stay in higher precision. AWQ \citep{lin2024awq} scales salient weight channels prior to quantization to protect them from rounding error; SqueezeLLM \citep{kim2024squeezellm} uses sensitivity-based non-uniform clustering of weights; SpQR \citep{dettmers2024spqr} keeps a small number of outlier weights at higher precision and quantizes the rest more aggressively; OmniQuant \citep{shao2024omniquant} learns clipping ranges and weight transformations end-to-end on calibration data; AQLM \citep{egiazarian2024aqlm} uses additive vector quantization for sub-3-bit weight compression. These methods are largely orthogonal to REAL-Q: they design new representations or optimize auxiliary quantization parameters (scales/shifts), while REAL-Q targets the sequential weight-updating solver itself. REAL-Q could in principle be combined with any of these representations.

\paragraph{Incoherence and rotation-based methods.} A separate family applies orthogonal transforms to the weight and activation matrices before quantization to reduce outlier magnitudes and produce more uniform distributions. QuIP \citep{chee2023quip} introduces the incoherence-processing framework with random orthogonal matrices and adaptive rounding; QuIP\# \citep{tseng2024quipsharp} extends this with structured (Hadamard) rotations and lattice codebooks. SliceGPT \citep{ashkboos2024slicegpt} uses orthogonal transformations to enable post-training structural pruning by safely reducing model dimensionality. Furthermore, methods like SpinQuant \citep{liu2024spinquant} and FlatQuant \citep{sun2025flatquant} actively learn rotation matrices or affine transformations to optimize weight and activation distributions. These pre-processing steps are complementary to the OBS-style update used by REAL-Q---we adopt QuaRot \citep{ashkboos2024quarot} as a default rotation pre-processing in all experiments.

\paragraph{Quantization-aware training and fine-tuning.} A different class of methods trades calibration efficiency for additional training: LLM-QAT \citep{liu2024llmqat} performs full quantization-aware training with distillation; LR-QAT \citep{bondarenko2024lrqat} reduces the cost of QAT via low-rank adapters; EfficientQAT \citep{chen2025efficientqat} uses block-wise alternating optimization of weights and quantization parameters; PV-tuning \citep{malinovskii2024pvtuning} fine-tunes the discrete codebook used by extreme low-bit quantization methods. These methods generally achieve higher accuracy than PTQ at low bit-widths but require multiple full epochs through training data and substantially more compute (typically $10$--$1000\times$ that of PTQ). REAL-Q sits firmly on the PTQ side of this trade-off: it incorporates first-order gradient corrections entirely within the single-pass GPTQ calibration pipeline, never updating dequantized weights globally or requiring full training epochs.

\paragraph{Hessian and Fisher information in compression.}
Second-order information has been extensively used for network compression. WoodFisher \citep{singh2020woodfisher} demonstrated the effectiveness of using the empirical Fisher matrix as a tractable proxy for the exact Hessian in OBS-based structural pruning. Similarly, \citet{liu2021groupfisher} derives Fisher-based metrics for structured pruning. HAWQ \citep{dong2019hawq} and HAWQ-V2 \citep{dong2020hawqv2} use Hutchinson's algorithm to efficiently estimate Hessian eigenvalues and traces for automated mixed-precision quantization. K-FAC \citep{martens2015kfac} approximates the Fisher matrix as Kronecker products for efficient natural gradient computation \citep{amari1998natural}.

\paragraph{Comparison with BRECQ.}
BRECQ \citep{li2021brecq} was originally proposed for CNN quantization, using a block-output Fisher-weighted reconstruction loss that is conceptually similar to our per-transformer-block Fisher MSE. However, BRECQ operates in a fundamentally different continuous optimization regime: it jointly optimizes all weight parameters of a residual block as continuous variables with a regularization penalty, rather than operating within a discrete, sequential GPTQ-style pipeline. Moreover, even if BRECQ's token-by-token diagonal Fisher objective were directly adapted for a transformer architecture, it would reduce to the same saliency-weighted Hessian approximation employed by GuidedQuant. To make this reduction explicit, BRECQ's per-token diagonal-Fisher reconstruction loss for a linear module $\hat{\bW}$ takes the form
\begin{equation}
    \cL_{\text{BRECQ}} = \frac{1}{T}\sum_{t=1}^{T} (\Delta\bW \bx_t)^\top \mathrm{diag}(\bg_t \odot \bg_t) \, (\Delta\bW \bx_t),
    \label{eq:brecq_loss}
\end{equation}
where $\Delta\bW = \bW - \hat{\bW}$ and $\bg_t = \nabla_{\by_t} \cL_{\text{task}}$ is the per-token output gradient. Expanding the diagonal weighting and grouping terms by output row $i$,
\begin{equation}
    \cL_{\text{BRECQ}} = \frac{1}{T}\sum_{i=1}^{m} \sum_{t=1}^{T} g_{t,i}^2 \, (\Delta\bW_{i,:} \bx_t)^2 = \sum_{i=1}^{m} \Delta\bW_{i,:} \underbrace{\Bigl(\tfrac{1}{T}\sum_{t=1}^{T} g_{t,i}^2 \, \bx_t \bx_t^\top\Bigr)}_{=:\,\bH_i} \Delta\bW_{i,:}^\top,
\end{equation}
which is exactly a per-output-row, saliency-weighted Hessian quadratic. This is the $N_g \!\to\! m$ limit of GuidedQuant's grouped Hessian (Eq.~\ref{eq:saliency_hessian}); GuidedQuant simply pools rows into $N_g$ groups for tractability. The structural limitations we discuss for GuidedQuant therefore apply identically, and the gradient correlation of this objective is shown to be low in our experiments (see Figure~\ref{fig:cosine_body}).

\section{Theoretical Analysis with Proofs}
\label{app:theory}

\subsection{Proof of Decoupled Fisher Aggregation (Proposition~\ref{prop:fisher_decoupled})}
\label{app:fisher_proof}

Here we provide the exact mathematical derivation for the approximation error introduced by the mean-field decoupling in Proposition~\ref{prop:fisher_decoupled}.

Using the cyclic property of the trace operator and the linearity of expectation, the exact expected second-order penalty can be rewritten as:
\begin{equation}
    \EE_t\bigl[\Delta\by_t^\top \bH_t \Delta\by_t\bigr] = \EE_t\bigl[\text{Tr}(\Delta\by_t^\top \bH_t \Delta\by_t)\bigr] = \text{Tr}\bigl(\EE_t[\bH_t \Delta\by_t \Delta\by_t^\top]\bigr).
\end{equation}
Similarly, the decoupled expectation used in our surrogate formulation evaluates to:
\begin{equation}
    \EE_t\bigl[\Delta\by_t^\top\, \EE_t[\bH_t]\, \Delta\by_t\bigr] = \text{Tr}\bigl(\EE_t[\bH_t] \cdot \EE_t[\Delta\by_t \Delta\by_t^\top]\bigr).
\end{equation}
By utilizing the linearity of the trace operator and the fact that both $\bH_t$ and $\Delta\by_t \Delta\by_t^\top$ are symmetric, the trace of their product is equivalent to their Frobenius inner product. Consequently, the difference between the exact joint expectation and the product of their marginal expectations exactly amounts to the sum of their element-wise covariances:
\begin{equation}
\text{Tr}\bigl(\EE_t[\bH_t \Delta\by_t \Delta\by_t^\top]\bigr) - \text{Tr}\bigl(\EE_t[\bH_t] \cdot \EE_t[\Delta\by_t \Delta\by_t^\top]\bigr) = \sum_{i,j} \text{Cov}\bigl((\bH_t)_{ij}, (\Delta\by_t \Delta\by_t^\top)_{ij}\bigr).
\end{equation}
Combined with the standard Fisher--Hessian identity $\EE_t[\bH_t] \approx \bF$, the decoupled formulation directly yields our objective $\cL_{\text{Fisher}}$ in Eq.~\ref{eq:fisher_loss}. The mathematical gap between the exact expected penalty and the decoupled surrogate is exactly the trace covariance $\text{Tr}\bigl(\text{Cov}(\bH_t, \Delta\by_t \Delta\by_t^\top)\bigr)$, which we explicitly trade for global cross-channel structural fidelity as discussed in Section~\ref{sec:fisher_loss}. \hfill$\square$

\subsection{Error Accumulation in Column-wise Quantization}
\label{app:error_accu}

Consider a single linear layer with weight matrix $\bW \in \RR^{m \times n}$ and input $\bx \in \RR^n$. The standard layer-wise quantization objective is:
\begin{equation}
    \min_{\hat{\bW}} \; \EE_{\bx} \left[ \| \bW \bx - \hat{\bW} \bx \|^2 \right] = \min_{\hat{\bW}} \; \text{Tr}\left( (\bW - \hat{\bW}) \bH (\bW - \hat{\bW})^\top \right),
\end{equation}
where $\bH = \EE[\bx \bx^\top] \in \RR^{n \times n}$ is the input covariance (Hessian of the quadratic reconstruction loss).

\begin{theorem}[Reconstruction error of GPTQ; restated from \citet{frantar2022obc}]
\label{thm:gptq_error}
Let $\bW$ be quantized column-by-column with GPTQ using a fixed Hessian $\bH$. Let $\boldsymbol{\epsilon}_j \in \RR^m$ denote the per-column quantization residual at step $j$, defined as $\boldsymbol{\epsilon}_j = \bW^{(j)}_{:,j} - Q(\bW^{(j)}_{:,j})$. The total layer-wise reconstruction error equals $\cL_{\text{GPTQ}} = \sum_{j=1}^{n} \|\boldsymbol{\epsilon}_j\|_2^2 / [\bH^{-1}]_{jj}$, exactly under the fixed-Hessian objective.
\end{theorem}

\paragraph{Proof of Theorem~\ref{thm:gptq_error}.} 
This follows from the OBS framework \citep{hassibi1993optimal,frantar2022obc}. At step $j$, quantizing column $j$ introduces error $\boldsymbol{\epsilon}_j$ and the optimal compensation for remaining columns is $\delta \bW_{:,j+1:n} = -\boldsymbol{\epsilon}_j \cdot [\bH^{-1}]_{jj}^{-1} \cdot \bH^{-1}_{j,j+1:n}$. By the Schur complement structure of $\bH^{-1}$, each row independently incurs a loss increase of $\epsilon_{ij}^2 / [\bH^{-1}]_{jj}$. Summing over all $m$ rows and $n$ columns yields the exact closed-form expression. \hfill$\square$

\begin{lemma}[Static Approximation Penalty]
\label{lem:staleness}
Let $\bH_{\mathrm{ideal}}^{(j)}$ denote the true downstream-aware Hessian at step $j$, and $\Delta \bH^{(j)} = \bH_{\mathrm{ideal}}^{(j)} - \bH$ denote the gap of the cached static Hessian. In GPTQ, the weight update at step $j$ spans multiple columns via compensation, forming a rank-1 perturbation $\Delta \bW^{(j)} = \boldsymbol{\epsilon}_j \mathbf{v}_j^\top$, where $\mathbf{v}_j \in \RR^n$ contains the static Hessian dependencies. Assuming the structural drift makes this static compensation increasingly misaligned, such that the projection gap is lower-bounded by $\mathbf{v}_j^\top \Delta \bH^{(j)} \mathbf{v}_j \geq c \cdot (j / n) \cdot \|\bH\|_2$ for some constant $c > 0$, the expected cumulative uncompensated penalty scales as:
\begin{equation}
    \EE[\mathcal{E}] = \Omega(n \cdot \bar{\epsilon}^2 \cdot \|\bH\|_2),
\end{equation}
where $\bar{\epsilon}^2 = \EE[\|\boldsymbol{\epsilon}_j\|_2^2]$ is the expected squared residual norm.
\end{lemma}

\paragraph{Proof of Lemma~\ref{lem:staleness}.} 
At step $j$, quantizing column $j$ introduces an immediate residual $\boldsymbol{\epsilon}_j = \bW^{(j)}_{:,j} - Q(\bW^{(j)}_{:,j})$ (defined as in Theorem~\ref{thm:gptq_error}, with $\bW^{(j)}$ the weight matrix entering step $j$). Following the OBS framework \citep{frantar2022obc}, GPTQ applies a correlated update to the remaining unquantized columns to compensate. The exact step-wise weight perturbation (new minus old) is a rank-1 matrix $\Delta \bW^{(j)} = \boldsymbol{\epsilon}_j \mathbf{v}_j^\top$, where the compensation vector $\mathbf{v}_j \in \RR^n$ is strictly defined as:
\begin{equation}
    [\mathbf{v}_j]_k = 
    \begin{cases} 
        0 & \text{if } k < j \\
        -1 & \text{if } k = j \\
        - [\bH^{-1}]_{j, k} / [\bH^{-1}]_{jj} & \text{if } k > j
    \end{cases}
\end{equation}
Evaluating this exact GPTQ update under the true downstream landscape, the actual step-wise penalty gap introduced by the stale Hessian is the quadratic form:
\begin{equation}
    \xi_j = \text{Tr}\left( \Delta \bW^{(j)} \Delta \bH^{(j)} (\Delta \bW^{(j)})^\top \right) = \left( \mathbf{v}_j^\top \Delta \bH^{(j)} \mathbf{v}_j \right) \cdot \|\boldsymbol{\epsilon}_j\|_2^2.
\end{equation}

In deep neural networks with non-quadratic loss landscapes, the true local Hessian $\bH_{\mathrm{ideal}}^{(j)}$ deviates from the initial static Hessian $\bH$ proportional to the accumulated distance traversed in the weight space. As the column index $j$ progresses, the accumulated weight displacement grows monotonically. It is therefore reasonable to assume that the structural drift penalty along the compensation direction, $\mathbf{v}_j^\top \Delta \bH^{(j)} \mathbf{v}_j$, worsens at least linearly with the macroscopic quantization progress $j/n$.

Formalizing this motivated assumption ($\mathbf{v}_j^\top \Delta \bH^{(j)} \mathbf{v}_j \geq c \frac{j}{n} \|\bH\|_2$ for some $c>0$), the step-wise penalty gap becomes $\xi_j \geq c \frac{j}{n} \|\bH\|_2 \|\boldsymbol{\epsilon}_j\|_2^2$.

Recognizing that in sequential quantization, the magnitude of the local quantization residual $\|\boldsymbol{\epsilon}_j\|_2^2$ is primarily determined by local weight variance, it can be reasonably modeled as statistically independent of the macroscopic structural drift index $j$. Taking the expectation over the random quantization residuals, we can decouple the summation due to linearity:
\begin{equation}
    \EE\left[ \sum_{j=1}^{n} \xi_j \right] \geq c \cdot \|\bH\|_2 \sum_{j=1}^{n} \frac{j}{n} \EE\left[ \|\boldsymbol{\epsilon}_j\|_2^2 \right].
\end{equation}
Letting $\bar{\epsilon}^2 = \EE[\|\boldsymbol{\epsilon}_j\|_2^2]$ denote the expected squared residual norm, the cumulative expected penalty evaluates to:
\begin{equation}
    c \cdot \|\bH\|_2 \cdot \bar{\epsilon}^2 \sum_{j=1}^{n} \frac{j}{n} = c \cdot \|\bH\|_2 \cdot \bar{\epsilon}^2 \cdot \frac{n(n+1)}{2n}.
\end{equation}
Since $\frac{n+1}{2} \geq \frac{n}{2}$, the expected total cumulative penalty is lower-bounded by $\frac{c}{2} \cdot n \cdot \bar{\epsilon}^2 \cdot \|\bH\|_2$. This establishes the asymptotic lower bound $\EE[\mathcal{E}] = \Omega(n \cdot \bar{\epsilon}^2 \cdot \|\bH\|_2)$, showing that under the stated assumption, static analytical solvers cannot prevent compounding cross-column errors. \hfill$\square$

\subsection{Descent Condition Analysis under First-Order Updates}
\label{app:blockgd_bounds}

Prior globally-guided PTQ methods rely on structural approximations to maintain the tractability of analytical solvers. In this section, we analyze REAL-Q through the lens of continuous surrogate optimization theory, deriving the condition under which descending our surrogate guarantees a decrease in the true end-to-end objective.

\begin{theorem}[End-to-End Descent Condition]
\label{thm:exact_descent}
Assume the true end-to-end objective $J(\bW)$ is $\beta$-smooth. We update the trailing un-quantized weights using a gradient step driven by our surrogate objective: $\bW^{(+)} = \bW - \eta \bg_{\mathrm{surr}}$, where $\bg_{\mathrm{surr}} = \nabla \tilde{J}(\bW)$ ($\tilde{J}$ denotes the surrogate objective). To strictly guarantee descent on the true global loss ($J(\bW^{(+)}) < J(\bW)$), the cosine similarity ($\cos \theta$) between the surrogate gradient $\bg_{\mathrm{surr}}$ and the true end-to-end gradient $\nabla J(\bW)$ must satisfy the exact strict inequality:
\begin{equation}
    \cos \theta > \frac{\eta \beta}{2} \cdot \frac{\|\bg_{\mathrm{surr}}\|}{\|\nabla J(\bW)\|}.
\end{equation}
\end{theorem}

\paragraph{Proof of Theorem~\ref{thm:exact_descent}.}
We rely on the standard Descent Lemma for $\beta$-smooth functions, which establishes a strict quadratic upper bound on the objective value after a parameter update:
\begin{equation}
    J(\bW - \eta \bg_{\mathrm{surr}}) \leq J(\bW) - \eta \langle \nabla J(\bW), \bg_{\mathrm{surr}} \rangle + \frac{\eta^2 \beta}{2} \|\bg_{\mathrm{surr}}\|^2.
\end{equation}
To ensure strict descent, the true loss after the update must be less than the initial loss ($J(\bW - \eta \bg_{\mathrm{surr}}) < J(\bW)$). This requires the sum of the linear and quadratic terms to be strictly negative:
\begin{equation}
    -\eta \langle \nabla J(\bW), \bg_{\mathrm{surr}} \rangle + \frac{\eta^2 \beta}{2} \|\bg_{\mathrm{surr}}\|^2 < 0.
\end{equation}
Moving the inner product to the right side yields:
\begin{equation}
    \eta \langle \nabla J(\bW), \bg_{\mathrm{surr}} \rangle > \frac{\eta^2 \beta}{2} \|\bg_{\mathrm{surr}}\|^2.
\end{equation}
We express the inner product using the cosine similarity between the two gradient vectors: $\langle \nabla J(\bW), \bg_{\mathrm{surr}} \rangle = \|\nabla J(\bW)\| \|\bg_{\mathrm{surr}}\| \cos \theta$. Substituting this into the inequality:
\begin{equation}
    \eta \|\nabla J(\bW)\| \|\bg_{\mathrm{surr}}\| \cos \theta > \frac{\eta^2 \beta}{2} \|\bg_{\mathrm{surr}}\|^2.
\end{equation}
Since the learning rate $\eta > 0$ and assuming a non-zero surrogate gradient ($\|\bg_{\mathrm{surr}}\| > 0$), we divide both sides by $\eta \|\nabla J(\bW)\| \|\bg_{\mathrm{surr}}\|$ to obtain the exact necessary condition:
\begin{equation}
    \cos \theta > \frac{\eta \beta}{2} \cdot \frac{\|\bg_{\mathrm{surr}}\|}{\|\nabla J(\bW)\|}.
\end{equation}
This inequality follows directly from the standard descent lemma without further relaxations. As empirically observed in Figure~\ref{fig:cosine_body}, REAL-Q's full aggregated Fisher matrix yields high $\cos \theta$, satisfying this condition under the learning rates used in our experiments. \hfill$\square$

\section{Additional Experimental Details}
\label{app:exp_details}

\subsection{Full Setup (continuation of Section~\ref{sec:exp_setup})}
\label{app:exp_details_full}

\paragraph{Optimizer.} We use Adam with $\beta_1=0.9$, $\beta_2=0.999$, $\epsilon=10^{-8}$, with bias correction applied at each block step. The learning rate depends on model and bit-width and is reported in Table~\ref{tab:realq_lr}; the final transformer block always uses a separate learning rate (also reported in Table~\ref{tab:realq_lr}) and optimizes the true KL loss against the LM head directly with Adam. The learning rate is fixed across all column blocks within a transformer block.

\paragraph{Reverse-cosine layer-wise learning-rate schedule.} The base learning rate of transformer block $i$ (out of $L$ total transformer blocks) is set to
\begin{equation}
    \eta_i = \eta_{\text{base}} + (\eta_{\text{final}} - \eta_{\text{base}}) \cdot \sin\!\left(\frac{\pi}{2} \cdot \frac{i-1}{L-1}\right),
    \label{eq:reverse_cosine}
\end{equation}
which interpolates from $\eta_{\text{base}}$ at the first block up to $\eta_{\text{final}}$ at the last block. The motivation is that early transformer blocks have not yet accumulated upstream quantization error and therefore require less aggressive correction; using a smaller learning rate at early blocks improves stability and prevents gradient explosions. For activation-aware quantization, this schedule is \emph{disabled} (constant $\eta_{\text{base}}$ throughout): once activation quantization is introduced, error accumulates from the very first block, and using a small early-layer learning rate would let this error compound rather than be corrected.

\paragraph{Activation clipping (\texttt{a\_loss\_clip}) for small Qwen3 models.} On Qwen3-0.6B, Qwen3-1.7B, and Qwen3-4B we additionally enable an output-perturbation clipping (\texttt{a\_loss\_clip}=0.95) when computing the Fisher MSE loss: per-token output channels whose $\Delta\by$ magnitude exceeds the 95th percentile are scaled down via a detached scale factor (so the loss is rescaled by the same factor and the gradient is not zeroed out). This is necessary because smaller post-trained Qwen3 models exhibit a small number of channels with very large $\Delta\by$, which can dominate the Fisher MSE and let the Adam updates be hijacked by a few outlier directions. Note this is distinct from the Hessian-side ``saliency clipping'' described below.

\paragraph{Hessian-side saliency clipping.} To prevent extreme gradient outliers (which can be 10--12 orders of magnitude larger than the median in deep layers) from collapsing the Hessian to a near rank-1 matrix, we clip per-token saliency values to the 99th percentile before computing the saliency-weighted Hessian.

\paragraph{Note on perplexity for Qwen3 models.} On the post-trained Qwen3 models we occasionally observe quantized perplexities slightly \emph{below} the bf16 baseline. This is expected: post-training does not optimize perplexity, so the bf16 model is no longer at the PPL minimum, and any perturbation---including quantization---can move PPL in either direction. KL divergence is the primary metric in all cases since it directly measures how far the quantized model's output distribution has moved from the full-precision reference.

\paragraph{Pre-computation of the aggregated Fisher and last-transformer-block handling.} The aggregated Fisher $\bF^{(\ell)}$ for each transformer block is computed once via a single end-to-end backward pass over the calibration set before quantization begins, with on-the-fly outer-product accumulation (no per-token storage; $O(d^2)$ memory per block). The final transformer block uses the true KL loss against the LM head (no Fisher surrogate, since there are no downstream blocks to traverse) with a separate learning rate (reported in Table~\ref{tab:realq_lr}), still using Adam.

\paragraph{FSDP multi-GPU pipeline for large models.} For LLaMA-3.1-70B the full-precision model and per-layer Fisher matrices do not fit in a single GPU. We use an FSDP-sharded \citep{zhao2023pytorch} end-to-end backward across multiple GPUs (Stage 0) to compute and cache Fisher / saliency to disk; Stage 1 then loads the cache and performs the per-layer quantization with Block-GD on a sharded model.

\subsection{Full W4A16 Results (continuation of Section~\ref{sec:exp_w4a16})}
\label{app:exp_full}

Table~\ref{tab:w4a16_main} reports the full seven-model W4A16 sweep summarized in Section~\ref{sec:exp_w4a16}. The LLaMA-3.1-70B run uses 256 calibration samples (rather than 2048) for tractability of the 70B forward / backward; all other rows use 2048 samples. REAL-Q achieves the lowest KL on every model.

\begin{table}[htbp]
    \caption{Full seven-model W4A16 (per-row weight quantization) results. KL ($\times 10^{-2}$) and PPL are evaluated on WikiText-2. Downstream columns report zero-shot accuracy (\%) on ten tasks and their average. bf16 KL is omitted (the bf16 model is the reference).}
    \label{tab:w4a16_main}
    \centering
    \scriptsize
    \setlength{\tabcolsep}{1.6pt}
    \begin{tabular}{ll*{13}{c}}
        \toprule
        Model & Method & KL & PPL & ARC-C & ARC-E & BoolQ & C-Eval & Hella. & LAMB. & OBQA & PIQA & SIQA & Wino. & Avg. \\
        \midrule
        \multirow{6}{*}{LLaMA-3.1-8B}
             & bf16 & --- & 6.25 & 53.41 & 81.31 & 82.17 & 48.66 & 78.90 & 75.33 & 44.80 & 81.18 & 50.15 & 73.80 & 66.97 \\
             & RTN & 16.1 & 7.64 & 49.23 & 76.81 & 79.24 & 38.86 & 75.08 & 70.33 & 43.60 & 78.07 & 49.85 & 71.67 & 63.27 \\
             & GPTQ & 4.95 & 6.60 & 53.75 & 80.01 & 81.53 & 46.51 & \textbf{78.20} & \textbf{75.02} & 44.20 & 80.58 & 49.59 & 72.45 & 66.18 \\
             & GPTAQ & 3.98 & 6.53 & 53.16 & \textbf{80.60} & 79.82 & 45.99 & 78.11 & 74.46 & 43.80 & 80.20 & 50.41 & \textbf{73.09} & 65.96 \\
             & GuidedQuant & 3.67 & 6.50 & 53.07 & 79.97 & \textbf{81.74} & \textbf{46.95} & 77.96 & 74.35 & 44.20 & 80.47 & 49.69 & 72.85 & 66.12 \\
             \rowcolor{gray!15}& REAL-Q (ours) & \textbf{3.36} & \textbf{6.48} & \textbf{53.84} & 79.97 & 81.07 & 46.36 & 78.08 & 74.91 & \textbf{46.20} & \textbf{80.96} & \textbf{50.61} & 72.85 & \textbf{66.48} \\
        \midrule
        \multirow{6}{*}{LLaMA-3.1-70B}
             & bf16 & --- & 2.81 & 64.85 & 86.53 & 85.44 & 63.60 & 85.06 & 78.91 & 48.20 & 84.33 & 54.86 & 79.56 & 73.13 \\
             & RTN & 71.0 & 6.83 & 51.79 & 76.47 & 78.29 & 31.43 & 75.48 & 63.05 & 40.20 & 79.33 & 43.71 & 71.35 & 61.11 \\
             & GPTQ & 18.7 & 3.43 & 63.31 & 86.32 & 85.50 & \textbf{64.64} & 84.50 & 78.40 & 47.60 & \textbf{84.22} & 49.74 & \textbf{80.03} & 72.43 \\
             & GPTAQ & 16.0 & 3.32 & 62.12 & 86.20 & 84.68 & 63.97 & \textbf{84.51} & 78.38 & 46.80 & 83.51 & 49.64 & 79.16 & 71.90 \\
             & GuidedQuant & 15.1 & 3.31 & \textbf{64.42} & 86.41 & \textbf{85.54} & 63.60 & 84.45 & \textbf{78.67} & 47.40 & 84.06 & \textbf{50.82} & 79.87 & \textbf{72.52} \\
             \rowcolor{gray!15}& REAL-Q (ours) & \textbf{14.3} & \textbf{3.30} & 63.91 & \textbf{86.49} & 84.83 & 62.78 & \textbf{84.51} & 78.44 & \textbf{47.80} & 83.79 & 49.23 & 79.56 & 72.13 \\
        \midrule
        \multirow{6}{*}{Qwen3-0.6B}
             & bf16 & --- & 20.95 & 33.70 & 55.72 & 63.91 & 43.76 & 47.28 & 40.03 & 31.20 & 67.63 & 43.30 & 55.96 & 48.25 \\
             & RTN & 55.5 & 34.37 & 28.75 & 41.96 & 57.68 & 31.58 & 43.19 & 23.64 & 29.60 & 63.11 & 41.91 & 53.51 & 41.49 \\
             & GPTQ & 30.2 & 28.37 & 29.01 & 49.49 & \textbf{68.93} & 29.64 & 42.61 & 30.95 & 29.40 & 64.85 & 42.89 & 55.01 & 44.28 \\
             & GPTAQ & 20.8 & 25.88 & 29.10 & 45.79 & 67.98 & \textbf{33.36} & 42.94 & 32.54 & 30.40 & 64.20 & 42.84 & 56.35 & 44.55 \\
             & GuidedQuant & 8.76 & 22.31 & 31.40 & 49.54 & 63.52 & 26.00 & 44.97 & 33.73 & \textbf{31.60} & 65.72 & \textbf{43.50} & 56.27 & \textbf{44.62} \\
             \rowcolor{gray!15}& REAL-Q (ours) & \textbf{6.79} & \textbf{21.57} & \textbf{32.08} & \textbf{50.63} & 56.54 & \textbf{33.36} & \textbf{45.18} & \textbf{34.91} & 30.80 & \textbf{66.00} & 38.74 & \textbf{56.83} & 44.51 \\
        \midrule
        \multirow{6}{*}{Qwen3-1.7B}
             & bf16 & --- & 16.73 & 42.75 & 69.91 & 77.95 & 58.62 & 60.37 & 50.44 & 36.80 & 72.47 & 44.22 & 60.93 & 57.45 \\
             & RTN & 120 & 75.61 & 30.72 & 37.67 & 64.80 & 42.20 & 49.67 & 24.49 & 29.20 & 63.00 & 41.56 & 54.06 & 43.74 \\
             & GPTQ & 21.3 & 21.50 & 34.13 & 56.36 & 76.91 & 52.23 & 56.94 & 42.60 & 34.40 & 67.74 & 43.30 & 57.46 & 52.21 \\
             & GPTAQ & 13.2 & 18.59 & 37.71 & 60.35 & 75.54 & 51.11 & 57.35 & 46.09 & 34.00 & 70.35 & 42.73 & 59.67 & 53.49 \\
             & GuidedQuant & 11.9 & 17.72 & \textbf{39.68} & \textbf{65.87} & 76.39 & 48.51 & 58.23 & 41.70 & \textbf{37.60} & \textbf{70.40} & \textbf{43.96} & \textbf{60.06} & 54.24 \\
             \rowcolor{gray!15}& REAL-Q (ours) & \textbf{6.07} & \textbf{16.54} & 38.40 & 58.67 & \textbf{77.98} & \textbf{55.42} & \textbf{58.67} & \textbf{48.22} & 36.20 & 70.29 & 42.84 & 59.75 & \textbf{54.64} \\
        \midrule
        \multirow{6}{*}{Qwen3-4B}
             & bf16 & --- & 13.66 & 54.01 & 78.41 & 85.20 & 70.43 & 68.40 & 59.29 & 40.40 & 75.03 & 51.28 & 66.14 & 64.86 \\
             & RTN & 31.1 & 17.11 & 46.76 & 69.82 & 80.46 & 62.85 & 64.80 & 50.42 & 38.40 & 73.29 & 48.41 & 62.67 & 59.79 \\
             & GPTQ & 13.5 & 13.58 & \textbf{50.26} & 74.20 & 83.49 & 61.89 & 66.67 & 59.07 & \textbf{40.20} & 74.70 & 48.46 & 63.93 & 62.29 \\
             & GPTAQ & 11.2 & 14.59 & 48.89 & 73.61 & \textbf{83.58} & 64.12 & 65.87 & 58.45 & 39.60 & 74.92 & \textbf{49.33} & 62.90 & 62.13 \\
             & GuidedQuant & 9.50 & 14.50 & 49.23 & 74.16 & 82.75 & 64.49 & 66.55 & 57.31 & 38.40 & 74.43 & 48.62 & 62.35 & 61.83 \\
             \rowcolor{gray!15}& REAL-Q (ours) & \textbf{5.44} & \textbf{13.44} & 49.74 & \textbf{74.62} & 83.46 & \textbf{67.38} & \textbf{67.06} & \textbf{60.12} & 39.40 & \textbf{75.90} & 46.32 & \textbf{65.11} & \textbf{62.91} \\
        \midrule
        \multirow{6}{*}{Qwen3-8B}
             & bf16 & --- & 9.74 & 56.23 & 80.72 & 86.51 & 79.42 & 74.73 & 64.35 & 41.80 & 77.91 & 52.56 & 68.03 & 68.23 \\
             & RTN & 19.6 & 10.85 & 51.62 & 76.26 & 84.53 & 73.25 & 71.50 & 63.26 & 39.60 & 75.84 & 49.18 & 67.32 & 65.24 \\
             & GPTQ & 3.96 & 9.88 & \textbf{55.46} & 79.50 & 86.57 & 77.71 & 73.79 & 62.82 & \textbf{40.60} & 76.71 & 51.74 & \textbf{68.19} & 67.31 \\
             & GPTAQ & 3.80 & 10.01 & 54.27 & 79.12 & \textbf{86.61} & 77.19 & 73.13 & 62.86 & 40.40 & 76.77 & \textbf{52.76} & 67.17 & 67.03 \\
             & GuidedQuant & 3.52 & 9.92 & 55.38 & 78.91 & 86.51 & \textbf{78.01} & 73.81 & \textbf{64.45} & 40.00 & \textbf{77.37} & 51.69 & 67.01 & 67.31 \\
             \rowcolor{gray!15}& REAL-Q (ours) & \textbf{3.26} & \textbf{9.76} & 54.52 & \textbf{80.64} & 86.45 & 77.41 & \textbf{73.88} & 63.73 & 40.20 & 77.20 & 51.59 & 68.11 & \textbf{67.37} \\
        \midrule
        \multirow{6}{*}{Qwen3-32B}
             & bf16 & --- & 7.61 & 60.75 & 83.38 & 86.82 & 85.96 & 82.58 & 67.24 & 46.20 & 81.94 & 54.04 & 72.93 & 72.18 \\
             & RTN & 21.8 & 8.57 & 58.02 & 77.31 & 86.67 & 82.54 & 80.30 & 64.45 & 47.40 & 79.54 & 50.82 & 71.90 & 69.90 \\
             & GPTQ & 4.98 & 7.74 & 58.87 & 81.61 & 86.39 & \textbf{85.14} & \textbf{82.32} & 67.05 & 46.20 & 81.56 & 53.12 & 72.14 & 71.44 \\
             & GPTAQ & 6.62 & 7.84 & 60.92 & 83.38 & 86.45 & 84.84 & 81.61 & \textbf{67.88} & 45.00 & 81.07 & 53.53 & \textbf{73.56} & 71.82 \\
             & GuidedQuant & 4.66 & 7.74 & \textbf{62.29} & \textbf{83.42} & 83.88 & 84.47 & 82.15 & 67.71 & \textbf{48.20} & \textbf{81.61} & \textbf{54.20} & 72.30 & \textbf{72.02} \\
             \rowcolor{gray!15}& REAL-Q (ours) & \textbf{4.24} & \textbf{7.53} & 60.84 & 83.29 & \textbf{87.43} & 84.92 & 82.19 & 67.34 & 45.40 & 81.39 & 51.84 & 72.38 & 71.70 \\
        \bottomrule
    \end{tabular}
\end{table}

\subsection{REAL-Q Learning Rates}
\label{app:realq_lr}

Table~\ref{tab:realq_lr} reports the non-final-block learning-rate hyperparameter and the final-transformer-block learning rate for every REAL-Q run. For rows using the reverse-cosine schedule, the reported value is $\eta_{\text{final}}$, and the first-block learning rate is fixed to $0.01\eta_{\text{final}}$ before layer-wise interpolation. For rows with the schedule disabled, the reported value is used as a constant learning rate throughout the non-final transformer blocks.

\paragraph{LR selection.} The per-model LRs in Table~\ref{tab:realq_lr} were obtained from a coarse log-scale grid search over $\{10^{-3}, 10^{-4}, \ldots, 10^{-7}\}$, conducted on a low-cost proxy of the full pipeline that uses a larger column block size and fewer backward samples per gradient step. We did not perform fine-grained per-model tuning; the LRs in Table~\ref{tab:realq_lr} are simply the best-performing point on this loose grid. Even with this coarse, low-fidelity LR sweep, REAL-Q already attains the substantial improvements over prior methods reported in Section~\ref{sec:experiments}, and we expect more careful per-model LR selection to only further widen the margin.

\begin{table}[htbp]
    \caption{REAL-Q learning rates used across experimental settings. The scheduled final LR is $\eta_{\text{final}}$ for reverse-cosine rows and the constant non-final-block LR for rows with the schedule disabled. The final-layer LR is used only for the final transformer block, where REAL-Q optimizes the true KL loss against the LM head.}
    \label{tab:realq_lr}
    \centering
    \small
    \setlength{\tabcolsep}{6pt}
    \begin{tabular}{llcc}
        \toprule
        Setting & Model / variant & Scheduled final / const.\ LR & Final-layer LR \\
        \midrule
        \multirow{7}{*}{W4A16}
            & LLaMA-3.1-8B & $1.0\!\times\!10^{-5}$ & $1.0\!\times\!10^{-6}$ \\
            & LLaMA-3.1-70B & $3.0\!\times\!10^{-6}$ & $1.0\!\times\!10^{-6}$ \\
            & Qwen3-0.6B & $3.0\!\times\!10^{-4}$ & $1.0\!\times\!10^{-5}$ \\
            & Qwen3-1.7B & $5.0\!\times\!10^{-4}$ & $1.0\!\times\!10^{-5}$ \\
            & Qwen3-4B & $2.0\!\times\!10^{-4}$ & $1.0\!\times\!10^{-5}$ \\
            & Qwen3-8B & $1.0\!\times\!10^{-6}$ & $1.0\!\times\!10^{-6}$ \\
            & Qwen3-32B & $2.0\!\times\!10^{-6}$ & $1.0\!\times\!10^{-6}$ \\
        \midrule
        W3A16 & Qwen3-8B & $2.0\!\times\!10^{-6}$ & $1.0\!\times\!10^{-6}$ \\
        W2A16 & Qwen3-8B & $3.0\!\times\!10^{-5}$ & $1.0\!\times\!10^{-6}$ \\
        \midrule
        \multirow{2}{*}{W4A4KV4}
            & Qwen3-8B (a-unaware) & $1.0\!\times\!10^{-6}$ & $1.0\!\times\!10^{-6}$ \\
            & Qwen3-8B (a-aware) & $1.0\!\times\!10^{-6}$ & $1.0\!\times\!10^{-6}$ \\
        \midrule
        \multirow{2}{*}{W3A4KV4}
            & Qwen3-8B (a-unaware) & $2.0\!\times\!10^{-6}$ & $1.0\!\times\!10^{-6}$ \\
            & Qwen3-8B (a-aware) & $3.0\!\times\!10^{-7}$ & $1.0\!\times\!10^{-6}$ \\
        \midrule
        \multirow{2}{*}{W2A4KV4}
            & Qwen3-8B (a-unaware) & $3.0\!\times\!10^{-5}$ & $1.0\!\times\!10^{-6}$ \\
            & Qwen3-8B (a-aware) & $5.0\!\times\!10^{-6}$ & $1.0\!\times\!10^{-6}$ \\
        \bottomrule
    \end{tabular}
\end{table}

\subsection{Experimental Details for Loss Approximation Quality}
\label{app:cosine_exp}

This section provides the detailed setup for the gradient alignment evaluation presented in Section~\ref{sec:cosine_body} (Figure~\ref{fig:cosine_body}).

We evaluate on Qwen3-0.6B at RTN W4A16. The model is processed transformer block by transformer block in the standard sequential quantization order. Immediately after the internal linear modules of block $\ell$ are quantized, we freeze the current model state to compute the exact analytical gradients. Specifically, for each linear module within block $\ell$, we calculate:
\begin{enumerate}
    \item[(i)] The \textbf{true end-to-end KL gradient} with respect to the module's weights, obtained via a full backward pass through all subsequent layers and the language modeling head over the calibration mini-batch.
    \item[(ii)] The \textbf{surrogate gradients} produced by the respective proxy losses: \textbf{REAL-Q Fisher MSE} (our aggregated full Fisher), \textbf{Saliency-guided MSE} (the saliency-weighted per-module MSE optimized by GuidedQuant), and \textbf{Module MSE} (the uniform per-module reconstruction MSE optimized by GPTQ / GPTAQ).
\end{enumerate}

We then compute the cosine similarity between the surrogate gradients and the true KL gradient. As observed in the per-module breakdown in the main text (Figure~\ref{fig:cosine_body}), while REAL-Q maintains strong overall alignment, its fidelity degrades slightly on FFN modules (particularly \texttt{up\_proj} and \texttt{down\_proj}) compared to attention modules. This is primarily due to the SwiGLU non-linearity, which makes the block-wise truncated second-order expansion a locally poorer approximation of the true downstream loss. This architectural heterogeneity directly motivates Limitation~(1) discussed in Appendix~\ref{app:limitations}.

\subsection{GPU-hour Overhead}
\label{app:gpu_hours}

All experiments were conducted using NVIDIA RTX Pro 6000 GPUs (96GB). Table~\ref{tab:gpu_hours} reports the total end-to-end wall-clock cost for applying REAL-Q at W4A16. This duration sums both the Stage-0 full-precision backward pass (for Fisher and saliency pre-computation) and the Stage-1 sequential quantization loop with Block-GD. 

\begin{table}[H]
    \caption{RTX Pro 6000 end-to-end REAL-Q W4A16 wall-clock per model. ``\#GPUs'' denotes the number of GPUs used for the run; ``Hours per GPU'' is the per-GPU wall-clock time; and ``GPU-hours'' represents the total compute budget consumed by the entire REAL-Q pipeline.}
    \label{tab:gpu_hours}
    \centering
    \small
    \begin{tabular}{lccc}
        \toprule
        Model & \#GPUs & Hours per GPU & GPU-hours \\
        \midrule
        LLaMA-3.1-8B  & 4 & 2.1  & 8.4   \\
        LLaMA-3.1-70B & 8 & 11.1 & 88.8  \\
        Qwen3-0.6B    & 4 & 0.18 & 0.72  \\
        Qwen3-1.7B    & 4 & 0.5  & 2.0   \\
        Qwen3-4B      & 4 & 1.2  & 4.8   \\
        Qwen3-8B      & 4 & 2.2  & 8.8   \\
        Qwen3-32B     & 8 & 7.4  & 59.2  \\
        \bottomrule
    \end{tabular}
\end{table}

While the inclusion of dynamic gradient corrections inherently introduces additional computational overhead compared to purely static, closed-form PTQ baselines like vanilla GPTQ, the dominant Stage-1 cost scales cleanly with the number of (transformer block $\times$ linear module $\times$ column block) triples. Consequently, the overhead grows roughly linearly with model size at a fixed block size $B$. Most importantly, the total compute budget remains comfortably within the practical realm of offline PTQ---requiring only a few GPU-hours for medium-sized models---which is orders of magnitude more efficient than quantization-aware training (QAT) or fine-tuning approaches that demand multiple full epochs over the training data.

\subsection{GPU Memory Footprint for Fisher Aggregation}
\label{app:memory_usage}

While REAL-Q preserves the full cross-channel structure of the Fisher (no row grouping), caching the full empirical Fisher Information Matrix ($\bF \in \RR^{d \times d}$) for all transformer blocks requires additional memory footprint during the offline calibration stage. 

Since the Fisher matrix is structured per transformer block and decoupled from the context length, the total memory required to store the aggregated statistics for the entire model is deterministic. Specifically, for a model with $L$ transformer blocks and a hidden dimension $d$, the storage cost using \texttt{bfloat16} precision (2 bytes per parameter) is exactly $2 \times L \times d^2$ bytes.

In Table~\ref{tab:fisher_memory}, we detail the theoretical memory footprint required to cache the full Fisher matrices for all seven models evaluated in our experiments. 

\begin{table}[h]
\centering
\caption{GPU Memory usage for storing full Fisher Information Matrices across different models (in \texttt{bfloat16} precision). The total memory exactly scales with $2 \times L \times d^2$.}
\label{tab:fisher_memory}
\resizebox{0.95\textwidth}{!}{
\begin{tabular}{lcccr}
\toprule
\textbf{Model Family} & \textbf{Blocks ($L$)} & \textbf{Hidden Dim ($d$)} & \textbf{Matrix Size per Block} & \textbf{Total Memory Cost} \\
\midrule
Qwen3-0.6B    & 28 & 1024  & $2.00$ MB  & $\sim 56$ MB \\
Qwen3-1.7B    & 28 & 2048  & $8.00$ MB  & $\sim 224$ MB \\
Qwen3-4B      & 36 & 2560  & $12.50$ MB & $\sim 450$ MB \\
Qwen3-8B      & 36 & 4096  & $32.00$ MB & $\sim 1.13$ GB \\
Qwen3-32B     & 64 & 5120  & $50.00$ MB & $\sim 3.13$ GB \\
\midrule
LLaMA-3.1-8B  & 32 & 4096  & $32.00$ MB & $\sim 1.00$ GB \\
LLaMA-3.1-70B & 80 & 8192  & $128.00$ MB& $\sim 10.00$ GB \\
\bottomrule
\end{tabular}
}
\end{table}

Regarding the transient memory required for the end-to-end backward pass in Stage 0, we treat it strictly as a standard distributed fine-tuning forward-backward step. By employing Fully Sharded Data Parallel (FSDP) combined with activation checkpointing across the calibration GPUs, the parameters, gradients, and optimizer states are sharded. This makes the per-GPU memory footprint for the calibration backward pass of massive models (e.g., LLaMA-3.1-70B) entirely scalable and consistent with standard LLM fine-tuning pipelines, requiring no exotic hardware beyond conventional multi-GPU clusters.

\subsection{Stability Across Random Seeds}
\label{app:seed_variance}

REAL-Q and all baselines are deterministic given a fixed calibration set, so the only stochastic source is the random sampling of WikiText-2 sequences used to form the calibration set. To assess the sensitivity of REAL-Q to this randomness, we re-run the full W4A16 pipeline on Qwen3-0.6B with five different calibration-sampling seeds while holding all other settings (model, bit-width, hyperparameters in Table~\ref{tab:realq_lr}, evaluation protocol) fixed. Seed~1 corresponds to the run reported in Tables~\ref{tab:w4a16_body} and~\ref{tab:w4a16_main}.

\begin{table}[h]
    \caption{REAL-Q W4A16 stability across five calibration-sampling seeds on Qwen3-0.6B. KL ($\times 10^{-2}$) and PPL are evaluated on WikiText-2. Seed~1 is the run reported in the main W4A16 tables. The reported standard deviation is the (sample) $1\sigma$ across the five runs.}
    \label{tab:seed_variance}
    \centering
    \small
    \setlength{\tabcolsep}{8pt}
    \begin{tabular}{lcc}
        \toprule
        Seed & KL ($\times 10^{-2}$) & PPL \\
        \midrule
        1 (reported) & 6.79 & 21.57 \\
        2 & 6.86 & 21.64 \\
        3 & 6.89 & 21.65 \\
        4 & 6.92 & 21.68 \\
        5 & 6.87 & 21.60 \\
        \midrule
        Mean $\pm$ std & $6.87 \pm 0.05$ & $21.63 \pm 0.04$ \\
        \bottomrule
    \end{tabular}
\end{table}

The standard deviation is roughly $0.7\%$ of the mean for KL and $0.2\%$ for PPL, indicating that the W4A16 result on Qwen3-0.6B is stable under calibration-sample randomness. Notably, even the worst of the five REAL-Q seeds (KL~$=6.92$, PPL~$=21.68$) remains substantially below the strongest prior baseline on this model (GuidedQuant: KL~$=8.76$, PPL~$=22.31$; cf.\ Table~\ref{tab:w4a16_main}), so the headline ranking is preserved across all seeds.

\section{Detailed Limitations and Future Directions}
\label{app:limitations}

The shift from static analytical solvers to dynamic gradient-based corrections in REAL-Q opens up several promising avenues for future research to further refine the methodology:

\textbf{(1) Enhancing Surrogate Fidelity for Severe Non-linearities.} 
As shown by the cosine alignment results in Appendix~\ref{app:cosine_exp}, while our aggregated Fisher MSE provides accurate gradient guidance for attention modules, its correlation with the true end-to-end gradient degrades on certain FFN linear modules (notably \texttt{up\_proj} and \texttt{down\_proj}). We attribute this to the SwiGLU non-linearity, which makes a pure second-order Taylor truncation a less faithful approximation of the downstream loss landscape. Future work could explore higher-order Taylor expansions or non-linear proxy functions specifically designed to capture FFN activation dynamics.

\textbf{(2) Cross-Module Optimizer State Transfer.} 
Currently, the Adam optimizer is re-initialized for each new linear module being quantized. Consequently, its first- and second-moment estimates have a very limited number of update steps to warm up before the module's quantization is finalized. Developing an optimizer with structured state-transfer mechanisms across linear modules---or designing custom optimizers tailored for short, non-stationary sequential trajectories---could significantly accelerate convergence and further improve REAL-Q's correction capabilities.

\textbf{(3) System-Level Memory Optimizations.} 
The Stage-0 pre-computation inherently requires temporarily keeping the full-precision model resident and storing per-layer Fisher matrices ($O(d^2)$ each). While we successfully mitigate this for 70B-class models using an FSDP two-stage pipeline, the peak memory overhead remains higher than that of vanilla, zero-shot GPTQ. We anticipate that this computational footprint can be systematically amortized in the future through custom memory-efficient backward kernels, intermediate activation offloading, and fused Block-GD operators, making the method even more accessible for consumer-grade hardware.


\end{document}